# Fragments-Expert: A Graphical User Interface MATLAB Toolbox for Classification of File Fragments

Mehdi Teimouri[*,1], Zahra Seyedghorban[1], Fatemeh Amirjani[1]

[1] Information Theory and Coding Laboratory, University of Tehran, Tehran, Iran.

* Corresponding author, e-mail: mehditeimouri@ut.ac.ir

**Abstract**

The classification of file fragments of various file formats is an essential task in various applications such as firewalls, intrusion detection systems, anti-viruses, web content filtering, and digital forensics. However, the community lacks a suitable software tool that can integrate major methods for feature extraction from file fragments and classification among various file formats. In this paper, we present Fragments-Expert that is a graphical user interface MATLAB toolbox for the classification of file fragments. It provides users with 22 categories of features extracted from file fragments. These features can be employed by 7 categories of machine learning algorithms for the task of classification among various file formats.

# 1 Introduction

The fastest way to identify file format is to check the file extension. This method is very unreliable and can be easily spoofed. Another method for file format identification is through magic bytes. Magic bytes are predefined sequences that exist in the file header. Most of the time, the examiner is dealing with a file fragment and not the whole file itself. So, checking the file extension or magic bytes is not helpful in practical situations.

In many real-world situations, the ability to identify the file type of a file fragment is necessary. File fragment classification plays an important role in firewalls, intrusion detection systems, anti-viruses, web content filtering, and digital forensics [1]. For example, in network intrusion detection, file fragment classification is used to filter out any suspicious data [2, 3]. Another practical application of file fragment classification is computer forensics and investigations. In this case, during the recovery and examination of given evidence, the examiner may deal with a collection of fragments with no available file system information [2, 4-9]. The third practical example is the file recovery process during which the file system structure is damaged or deleted [10-12].

Most of the research projects in this area have focused on content-based file fragment classification. Many research projects in content-based file fragment classification, extract some features such as byte frequency distribution (BFD), Shannon entropy, Kolmogorov complexity, longest common strings and longest common subsequences [9] [13]. Then, a decision machine model (e.g. naïve Bayes, *k*-nearest neighbors, decision tree, random forest, and support vector machine) is trained using these features. In older research projects, the researchers utilize feature threshold comparison schemes to make decisions about the type of the fragments [11, 14]. Usually, these threshold-based methods do not employ any learning algorithm.

Some content-based methods have employed clustering for their purpose. Nguyen et al. investigate the identification of image type data [15]. They cluster uncompressed chunk data into three categories based on the entropy measure. Then, they employ a decision tree based on byte frequency distribution for each category. Li et al. have introduced three algorithms based on centroid models: single-centroid (one model for each file type), multi-centroid (multiple models for each file type), and exemplar files (the set of files of each file type) [16]. The mean and standard deviation of the byte frequency distribution of each file type are considered as centroid models. Then, Mahalanobis and Manhattan distances are employed for identifying file types.

There is a variety of Commercial-Off-The-Shelf (COTS) software for file type identification. These solutions mostly perform as part of a forensics or file recovery tool. Some well-known COTS tools are LibMagic [17], TrID [18], Oracle Outside In Technology [19], DROID [20], JHOVE [21], Autospy [22] (which uses PhotoRec [23] for type detection), ExifTool [24], FTK [25], EnCase [26], AnalyzeIt [27], and Toolsley [28]. These software packages rely on signature bytes for file type detection. So, they are not suitable for file fragment classification. Gopal et al. evaluated the performance of several software packages. They reported that most solutions fail when signature bytes are missing [3]. Konstantinos conducted several experiments using some of these recovery tools. The result shows that most of them misclassify a file format when the signature bytes are altered [29].

Some tools have been developed for file fragment classification. Table 1 shows a summary of these methods. SÁDI employs a limited number of features and the classification approach is simple. This tool finds the minimum and maximum values of each feature for each file type [30]. Then, it checks to see if an instance falls in these ranges. NetFox is another toolbox that is implemented as part of a network forensics framework. A part of NetFox has the capability of detection of codecs within an RTP stream [31]. JPGcarve is a file carving tool specialized for the recovery of JPG images from hard drive data dumps. The method relies on cluster size and positions in the hard disk space [10]. It conducts a search space in the data dumps and uses entropy thresholding for identification. A relatively good open-source software is Sceadan. Sceadan uses SVM and a set of statistical features for identification. The authors have reported an accuracy of 73.4% across 38 different file types [32]. An almost new approach is FiFTy [33], which is an open-source tool for file type identification. FiFTy uses a compact neural network with an embedding space for automatic feature extraction. The authors have reported their results for two different fragment sizes and 75 file types.

As seen above, there is a little amount of publicly available computer codes in this area and most of them employ small feature sets and limited machine learning methods. In this paper, we present Fragments-Expert. Fragments-Expert is an open-source graphical user interface (GUI) MATLAB toolbox that tries to fill this void by providing the researchers with many feature extraction methods and machine learning algorithms in the field of file fragment classification. Fragments-Expert enables researchers to directly re-use our implementations for the task of file fragments classification. They can also make desirable changes to our implementations to test the effect of employing different and innovative features.

Table 1: Tools and software packages for file fragment classification.

| Tool | Year | Method and Features | Case study |
|---|---|---|---|
| SÁDI [30] | 2008 | **Method:** Determining ranges for statistics values for differentiation among different data types<br><br>**Features**: Average, kurtosis, distribution of averages, standard deviation, distribution of standard deviations, and byte distribution. | **Fragments size**: 256 bytes<br><br>**Data types**: BMP, CSV, DLL, EXE, HTML, JPG, TXT, and XLS |
| Sceadan [32] | 2013 | **Method:** Support Vector Machine<br><br>**Features**: Entropy, Kolmogorov complexity, mean, standard deviation, mean absolute deviation, Hamming weight, kurtosis, skewness, average contiguity between bytes, maximum byte streak, and low, medium, and high ASCII frequencies | **Fragment size**: 512 bytes<br><br>**Data types**: 38 data types |
| NetFox [31] | 2014 | **Method:** Identify codecs by comparing the feature set of a stream with the Codec Mapper Table structure<br><br>**Features**: specific features of RTP packets | **Fragment size**: Packet size of real RTP streams<br><br>**Data type**: RTP streams encoded using different codecs and sampling frequencies. |
| JPGcarve [10] | 2015 | **Method:** Simple heuristics and entropy estimation<br><br>**Features**: Cluster size, offset, start and end position of a fragment, and entropy | **Fragment size**: From 512 bytes to cluster size<br><br>**Data type**: JPEG |
| FiFTy [33] | 2019 | **Method:** Compact Neural Network<br><br>**Features**: Using the embedding layer to extract features. | **Fragment Size**: 512 and 4096 bytes<br><br>**Data type**: 75 data types |

The remainder of this paper is organized as follows. In Section 2, we introduce 22 categories of features that are widely used in literature to describe file fragments. The architecture of Fragments-Expert is presented in Section 3. For a user who is an expert in coding with MATLAB, this section provides insight for modifying the toolbox for further functionalities. On the other hand, this section can help a regular user to understand the structure of the toolbox.

In Section 4, the functionalities of the toolbox are presented. Generating fragments, feature extraction, and feature selection are all described in this section. The processes of training, test, and cross-validation with available machine learning methods are also explained in this section. Data visualizing tools are also covered in this section. Finally, the processes of loading previously generated datasets and results are explained in this section.

An illustrative example is provided in Section 5. This example demonstrates how the toolbox functionalities can be employed in a practical problem. Finally, in Section 6, program availability and limitations are discussed.

# 2 The Features Used in File Fragment Classification

In this section, we present the most common feature categories that are proposed in the literature. These features are included in Fragments-Expert. So, when we describe these features, our choices of implementations are also explained.

## 2-1 Byte Frequency Distribution

This feature type gives you a vector with length 260. The first 256 values are BFD values. These values show the normalized frequency of byte values 0 to 255. $BFD_i$ is the normalized frequency of byte value $i = 0, 1, \ldots, 255$,

$$BFD_i = p_i \times 256, \tag{1}$$

where $p_i$ is the estimated probability of each byte value that is calculated as

$$p_i = \frac{f_i}{L}, \tag{2}$$

in which $f_i$ is the frequency of $i$th byte value and $L$ is the length of the fragment in bytes.

The other four features computed along with BFD are as follows. The first three features are defined in [9].

- SdFreq: Standard deviation of byte frequencies
- ModesFreq: The sum of the four highest byte frequencies
- CorNextFreq: Correlation of the frequencies of byte values $j$ and $j+1$
- ChiSq: The chi-square goodness of the fit test is a measure of randomness. Here we are using this measurement to compare the observed distribution of byte values to a uniform random distribution. For each byte value, we calculate the sum of the squared differences between the observed frequency of the byte values in the fragment and the expected frequency of the byte values in a uniform random distribution divided by the expected frequencies as follows [34].

$$T = \sum_{i=0}^{n=1} \frac{(f_i - \frac{L}{256})^2}{\frac{L}{256}}. \tag{3}$$

Then we use `chi2cdf(T,255,'upper')` command to compute $p - value$ for this test.

## 2-2 Rate of Change

Rate of change (RoC) is the vector consist of the frequencies of the absolute values of the differences between consecutive byte values in a file fragment. If we denote a fragment of bytes with length $L$ by $\mathbf{x} = [x_1, x_2, \ldots, x_L]$, the $j$th element of RoC vector, $j = 1, 2, \ldots, L - 1$, is calculated as follows.

$$y_j = \frac{|x_i - x_{i+1}|}{L - 1}. \tag{4}$$

Then each element of the RoC vector is normalized as follows.

$$\hat{y}_j = \begin{cases} \dfrac{256 \times 256}{2 \times (256 - j)} \times y_j & j \neq 0 \\ 256 \times y_j & j = 0 \end{cases}. \tag{5}$$

When the user includes this feature type in the feature extraction process, the mean of the rate of changes is also calculated as a feature. So, in this case, the feature vector is a vector with length 257.

## 2-3     Longest Contiguous Streak of Repeating Bytes

This feature is the normalized length of the longest contiguous streak of repeating bytes in the fragment. For example, for vector

[12  123  123  123  43  123  43  43  43  43  43  123  43  76  54  54  54  54],

the byte value 43 is repeated consecutively five times. So, in this case, the output feature is 5/18.

## 2-4     $n$-grams

In this case, the frequencies of the bit sequences with length $n$ are calculated. This feature is defined for bitstreams. So, the fragment of bytes is first converted to its equivalent binary form. As an example, consider the binary vector [1 0 1 1 1 0 0 1 0 1 1 0 1 0 0 0] with length 16. To calculate 2-grams for this vector, first, all possible combinations of zeros and ones with length two must be counted. For example, the bit pattern 0 0 is repeated three times. So, the normalized 2-gram value for this pattern is equal to $\dfrac{3}{16-2+1} = 0.2$.

When you include this feature type in the feature extraction process, you must choose the $n$ values in the next window of GUI. You can enter a vector of integer values for $n$. In this case, $n$-grams for all the values in this vector will be calculated. In the current version, $n$ values can be any integer value from 1 to 13.

## 2-5     Byte Concentration Features

From the BFD values described in Section 2-1, three features are calculated [9]:

- Low: Sum of the $BFD_i$ values for $0 \leq i < 32$.
- ASCII: Sum of the $BFD_i$ values for $32 \leq i < 128$.
- High: Sum of the $BFD_i$ values for $192 \leq i < 256$.

## 2-6     Basic Lower-Order Statistics

By selecting this feature type, seven different statistics are calculated as follows. First, three different mean values, i.e. arithmetic, geometric, and harmonic, are calculated as follows.

$$\mu_A = \frac{1}{L} \sum_{i=1}^{L} x_i, \tag{6}$$

$$\mu_G = \left( \prod_{i=1}^{L} x_i \right)^{\frac{1}{L}}, \tag{7}$$

and

$$\mu_H = \frac{L}{\sum_{i=1}^{L} \frac{1}{x_i}}. \tag{8}$$

Moreover, the standard deviation (STD) is calculated as follows.

$$\sigma = \sqrt{\frac{1}{L-1} \sum_{i=1}^{L} |x_i - \mu_A|^2}. \tag{9}$$

The mode value, which is the most frequently occurring value in **x**, is also calculated. The median value, which is the median value among all values of **x**, is also calculated. Finally, the MAD value, which is the mean absolute deviation of values, is calculated as $mean\left(abs(\mathbf{x} - mean(\mathbf{x}))\right)$.

## 2-7     Higher-Order Statistics

The unbiased estimation of kurtosis and skewness is employed in Fragments-Expert. Kurtosis estimation is calculated as follows

$$kurtosis = \frac{L-1}{(L-2)(L-3)}\left((L+1)K - 3(L-1)\right) + 3, \tag{10}$$

where $K$ is

$$K = \frac{\frac{1}{L}\sum_{i=1}^{L}(x_i - \mu_A)^4}{(\frac{1}{L}\sum_{i=1}^{L}(x_i - \mu_A)^2)^2}. \tag{11}$$

Moreover, skewness estimation is calculated as follows

$$skewness = \frac{\sqrt{L(L-1)}}{L-2} S, \tag{12}$$

where $S$ is

$$S = \frac{\frac{1}{L}\sum_{i=1}^{L}(x_i - \mu_A)^3}{(\sqrt{\frac{1}{L}\sum_{i=1}^{L}(x_i - \mu_A)^2})^3}. \tag{13}$$

## 2-8     Bicoherence

Bicoherence is a higher-order statistic that measures the non-linearity and non-Gaussianity in a given fragment of byte values. Here we computed the average bicoherence according to [35].

## 2-9     Window-Based Statistics

These features are defined according to [36]. For calculating these features you must at first choose a window size $W$. Five statistics including delta moving average, delta$^2$ moving average, delta standard deviation, delta$^2$ standard deviation, and deviation from the standard deviation are obtained from the values in the file fragment using non-overlapping and consecutive windows.

The moving average is calculated by taking the average of the mean of byte values in all windows. If $\mu_j; j = 1, 2, \ldots, J = \left\lfloor \frac{L}{W} \right\rfloor$ denotes the average of byte values in window $j$, the delta moving Average is calculated as follows

$$\Delta \mu = \frac{1}{J-1} \sum_{j=1}^{J-1} |\mu_{j+1} - \mu_j|. \tag{14}$$

Moreover, delta² moving average is calculated as follows

$$\Delta\Delta \mu = \frac{1}{J-2} \sum_{j=1}^{J-2} \left||\mu_{j+2} - \mu_{j+1}| - |\mu_{j+1} - \mu_j|\right|. \tag{15}$$

Similarly, if $\sigma_j; j = 1, 2, \ldots, J$ denotes the standard deviation of byte values in window $j$, delta standard deviation, and delta² standard deviation are respectively calculated as follows

$$\Delta \sigma = \frac{1}{J-1} \sum_{j=1}^{J-1} |\sigma_{j+1} - \sigma_j|, \tag{16}$$

$$\Delta\Delta \sigma = \frac{1}{J-2} \sum_{j=1}^{J-2} \left||\sigma_{j+2} - \sigma_{j+1}| - |\sigma_{j+1} - \sigma_j|\right|. \tag{17}$$

Finally, the deviation from the standard deviation is calculated as follows

$$d_\sigma = \frac{1}{J} \sum_{i=1}^{J} |\sigma_j - \sigma|. \tag{18}$$

## 2-10 Autocorrelation

This feature is the sample autocorrelation of the input data fragment up to a specific lag value $\ell$. The sample autocorrelation measures the correlation between $x_t$ and $x_{t+k}$, where $k = 1, \ldots, \ell$. The sample autocorrelation for lag $k$ is

$$r_k = \frac{c_k}{c_0}, \tag{19}$$

where

$$c_k = \frac{1}{L-1} \sum_{j=1}^{L-k} (x_j - \mu_A)(x_{j+k} - \mu_A). \tag{20}$$

Note that $c_0 = \sigma^2$ is the sample variance of the fragment. By choosing this feature, you have to specify the maximum lag value in the next window of the GUI.

## 2-11 Frequency Domain Statistics

As defined in [13], the frequency spectrum of a vector (in or context, a fragment) can be divided into equal sub-bands. Then the features of mean, variance, and skewness are extracted for each sub-band. For example, if the number of sub-bands is equal to four, 12 frequency-domain statistics are extracted. You can set the number of sub-bands from 1 to 8.

## 2-12 Binary Ratio

The definition of Binary Ratio (BRO) is obtained from [14] and is calculated over the bitstream (i.e. the equivalent binary form of the fragment). Each byte value is converted to an 8-bit representation. If $\mathbf{b} = [b_1, b_2, \ldots, b_{8 \times L}]$ denotes the equivalent binary form of the fragment, the Binary Ratio is defined as follows.

$$\text{BRO} = \frac{\sum_{j=1}^{L}(1-b_i)}{\sum_{j=1}^{L} b_i}. \tag{21}$$

## 2-13    Entropy

In this case, two statistics are obtained for the fragment: entropy and the difference between *L*-truncated entropy of uniform distribution [37] and the value of obtained entropy.

Entropy in information theory is a measure of the uncertainty of a sequence. In our case, in which we have a fragment with length $L$, the Shannon entropy is defined as

$$H = -\sum_{i=0}^{255} p_i \log_2 p_i. \tag{22}$$

The *L*-truncated entropy of a uniform distribution can be approximated as [37]

$$H_L(U) \cong \log m + \log c - e^{-c} \sum_{j=1}^{+\infty} \frac{e^{j-1}}{(j-1)!} \log j, \tag{23}$$

where $m$ is the number of possible byte values (which, in this case, is 256) and $c = \frac{L}{m}$.

## 2-14    Video Patterns

Some video formats have repeating patterns in their payload. The occurrence rate of these patterns can be used to determine the format of a file fragment. Here, we count the occurrences of these patterns and to make the resulting number independent of the length of the fragment and also the length of the pattern, we normalize it as follows

$$\frac{frq}{L - l_p + 1} \times 2^{8 \times l_p}, \tag{24}$$

where $frq$ is the number of occurrences of the pattern in the fragment and $l_p$ is the length of the pattern in bytes. Here, 17 patterns are considered. These patterns correspond to five different video formats. So, in this case, the feature vector contains 17 normalized pattern frequencies. In Table 2, employed video patterns are presented. The patterns for MKV, AVI, RMVB, and MP4, are proposed in [38].

**Table 2: Different byte patterns for video formats presented in hexadecimal format.**

| Format | Pattern(s) |
|--------|------------|
| MKV | 0xA0 |
|  | 0xA3 |
| AVI | 0x30306463 |
|  | 0x30317762 |
| RMVB | 0x0000 |
|  | 0x0001 |
| OGV | 0x4F676753 |
| MP4 | 0x419A |
|  | 0x019E |
|  | 0x019F |
|  | 0x419B |
|  | 0x6742 |
|  | 0x419E |
|  | 0x419F |
|  | 0x6588 |
|  | 0x68CE |
|  | 0x6588 |

## 2-15    Audio Patterns

This feature type calculates the normalized frequencies of specific audio bit patterns (i.e. sync words) in a file fragment. First, each byte value is converted to an 8-bit representation. Then, similar to the calculation of video patterns, normalization is applied to make the count of the occurrences independent of the length of the fragment and also the length of the pattern. In Table 3, the employed audio patterns are presented.

**Table 3: Different bit patterns for audio formats [39].**

| Format | Pattern |
|--------|---------|
| MP3 | 1111 1111 1111 |
| FLAC | 1111 1111 1111 10 |

## 2-16    Kolmogorov Complexity

The Kolmogorov complexity of a fragment is a measure of the computational resources needed to specify the fragment. But, no general algorithm can determine this complexity. We employ the method of [40] for estimating this complexity. For the sake of run-time speed, this function is written in C-MEX format.

## 2-17  False Nearest Neighbors and Lyapunov Exponents

Two chaotic features false neighbors fraction (FNF) and Lyapunov exponent (LE) are implemented in Fragments-Expert. Hicsonmez et al employed chaotic features for the identification of audio codecs [13]. The concept of the chaotic features is based on the neighborhood of the signal vectors. Let's define the signal vector $\mathbf{s}_i = [x_i, x_{i+1}, \ldots, x_{i+(D-1)}]$, where $D$ is the embedding dimension of the phase space. The distance between two nearest neighbors is defined as

$$d_D(\mathbf{s}_i, \mathbf{s}_j) = \sqrt{\sum_{k=0}^{D-1}(x_{i+k} - x_{j+k})^2}, \tag{25}$$

where $\mathbf{s}_j$ is the nearest neighbor of $\mathbf{s}_i$ on a nearby trajectory. If $d_D(\mathbf{s}_i, \mathbf{s}_j)$ is significantly different from $d_{D+1}(\mathbf{s}_i, \mathbf{s}_j)$, then $\mathbf{s}_i$ and $\mathbf{s}_j$ are considered to be a pair of false neighbors. After labeling all neighbors as true or false, the FNF is defined as the ratio of false neighbors to all neighbors [13]. So, we can define the feature vector $F_D$ with three components: the fraction of false neighbors, the average size of the neighborhood, and the root mean squared (RMS) size of the neighborhood.

$$F_D = \{FNF, mean\left(d_D(\mathbf{s}_i, \mathbf{s}_j)\right), RMSE\left(d_D(\mathbf{s}_i, \mathbf{s}_j)\right)\} \tag{26}$$

The Lyapunov exponent is a chaotic feature that quantifies the predictability of a signal. A system with a greater magnitude of LE is said to be more unpredictable [13]. The LE is calculated for each embedding dimension $D$ as

$$\lambda_D = \lim_{K \to +\infty} \frac{1}{K} \sum_{i=1}^{K} \log \frac{d_D(\mathbf{s}_{i+1}, \mathbf{s}_{j+1})}{d_D(\mathbf{s}_i, \mathbf{s}_j)}. \tag{27}$$

When you choose this feature type, you have to provide three parameters: the ratio factor, which is used to determine the false neighbors, and the minimum and maximum values for embedding dimensions. For the sake of run-time speed, this function is written in C-MEX format.

## 2-18  GIST Features

Ming Xu et al expressed that a vector of byte values (i.e. a fragment) can be reshaped into a matrix and regarded as a grayscale image [41]. So, they used GIST Descriptor that performs very good in scene and object classification [42]. To transform a fragment to a grayscale image and extracting GIST features, some parameters have to be set: Image row size, the number of non-overlapping windows in each dimension, which we denote by $M$, and the number of orientations for each scale. The last parameter is a vector of integer values which we denote by $[O_1, O_2, \ldots]$. The result of this type of feature extraction is a feature vector with length $M^2 \sum_i O_i$

## 2-19  Longest Common Substrings and Longest Common Subsequences

We define the longest common substring of two fragments to be the longest byte pattern existing in both fragments. A subsequence of a byte stream is any sequence of bytes obtained by deletions from the original. Accordingly, we can define the longest common subsequence of two fragments.

To define the longest common substring and the longest common subsequence features, we need certain fragments, which are called representatives. For each fragment in the dataset, the longest common substring and the longest common subsequence between that fragment and each representative are calculated. When using these features you have to select which class you want the representatives to be obtained from and where in the dataset the representatives should be taken (i.e. from the beginning of dataset, end of the dataset, or random positions in the dataset). Also, you must input the number of representative samples for the selected class. Corresponding to each representative class, two features are computed for a fragment: the average length of the longest common substrings and the average length of the longest common subsequences. For the sake of run-time speed, this function is written in C-MEX format.

### 2-20    Centroid Models

Using a class of representatives, you can obtain a measure of similarity between the byte frequency distribution of that class instances and the investigated fragment. The value of similarity can then be used as a feature for classification.

In Fragments-Expert, we have implemented two similarity measures: cosine similarity and Mahalanobis distance. When you select this feature type, you are asked to select the classes of representatives for centroid models among all class labels. Then you must enter the number of representatives for each class and where in the dataset the representatives should be taken.

After the parameters are set, BFD of the representative instances is calculated. Then for each byte value of the BFD, the mean and the standard deviation values are calculated. This mean and standard deviation values form a centroid model. By denoting the mean and the standard deviation BFD values of the representative instances by $\mu_i^c$ and $\sigma_i^c$ ($i = 0, 1, \ldots, 255$), we calculate the similarity features for each investigated fragment using

$$\text{CosineSimilarity} = \frac{\sum_i (BFD_i \times \mu_i^c)}{\sqrt{\sum_i BFD_i^2} \times \sqrt{\sum_i \mu_i^{c\,2}}} \tag{28}$$

and

$$\text{MahalanobisDistance} = \sqrt{\sum_i \frac{(BFD_i - \mu_i^c)^2}{(0.01 + \sigma_i^c)}}, \tag{29}$$

where $BFD_i$ is the normalized frequency of byte values $i = 0, 1, \ldots, 255$ in the examined fragment.

# 3  Toolbox Architecture

Figure 1 shows the four main modules of Fragments-Expert: Fragments Generation, Feature Extraction/Selection, Classification of Fragments, and Data Visualization. In each module, multiple tools are available that enable the researchers to simply examine the effect of different feature types and classification methods.

- **Fragments Generation:** Using Fragments-Expert, you can generate fragments with specific parameters from raw multimedia files. This procedure is explained in Section 4-1.

- **Feature Extraction/Selection:** Most common feature types are available in Fragments-Expert. In Section 4-2, generating a dataset of features is presented. The details of these features are discussed in Section 2. In Section 4-3, generating a dataset for an already trained decision machine is discussed. Two common feature selection algorithms are also implemented in Fragments-Experts that are described in Section 4-4. Various operations on datasets (such as randomization of Datasets) are also available in Fragments-Expert. These operations are explained in Section 4-5.

- **Classification of Fragments:** To classify file fragments into file formats, the user can employ various common machine learning methods. Training, test, and cross-validation are provided for each method. The processes for training, test, and cross-validation are explained in Sections 4-6, 4-7, and 4-8, respectively.

- **Data Visualization:** This module offers some tools for visualizing data samples. Visualization tools are explained in Section 4-9.

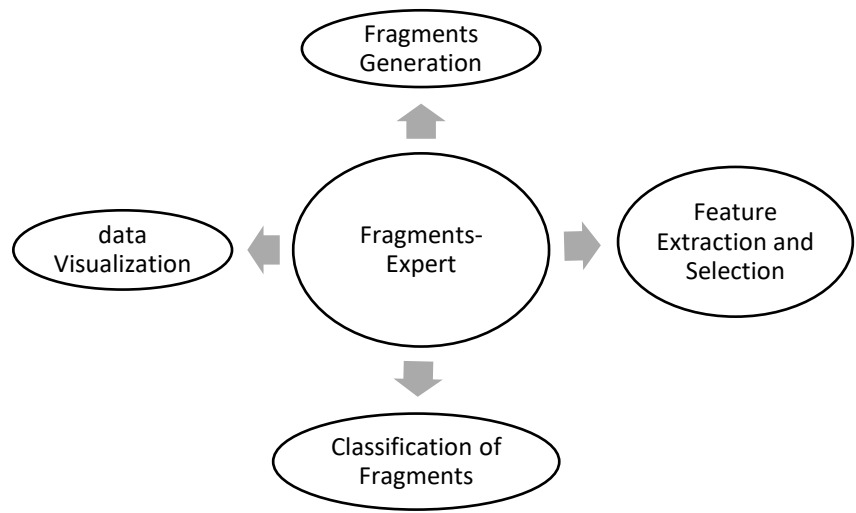

Figure 1: Four main modules of Fragments-Expert.

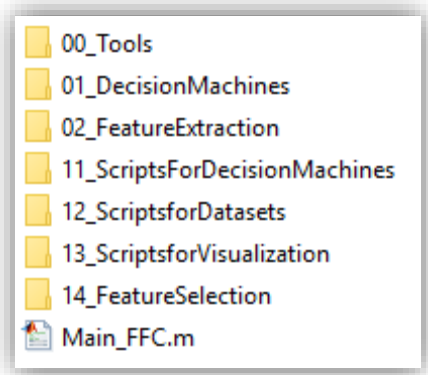

Figure 2: The structure of the source folder of Fragments-Expert.

As can be seen in Figure 2, the toolbox source folder includes seven subfolders and an m-file named `Main_FFC.m`. By running `Main_FFC`, the GUI of the toolbox will open. All the toolbox functionalities are available through this GUI. These functionalities are discussed in Section 4. A regular user needs only to interact with this GUI.

The seven subfolders in the toolbox source folder are as below. Understanding the structure of these subfolders helps an expert-user in MATLAB to modify the toolbox for further functionalities.

- **00_Tools**: This subfolder contains various tools. These tools are as follows.
    - A class of these tools, which are called "Dataset Tools", are the tools that provide functionalities of loading, expanding, random permutation, and partitioning datasets. Also, the tools for merging labels in a dataset or extracting a sub-dataset from a dataset are among these tools. Another tool is a tool for scaling the features in a dataset to a range.
    - Another class of these tools, which are called "Decision Machines Tools", are the tools that provide functionalities of loading decision machines and also the test and cross-validation results of decision machines. Two other tools in this class of tools are the tools used for constructing the confusion matrix of classification results and also scaling the rows of confusion matrix to percent values. Another tool is a tool that assigns weights to the samples of a dataset according to a weighting strategy.
    - Another class of the tools is "Files and Fragments Tools". This class of tools helps to determine the number of subfolders and files in a given folder, together with their names and sizes. Also, a useful tool, which is written as an m-file named `TakeRandomFragments_FFC`, enables the functionality of extracting random fragments from a file. Another useful tool, which is written as an m-file named `SetVariableNames_FFC`, provides a GUI to guide the user to assign valid MATLAB variable names to a group of labels.
    - Another class of the tools, which are called "User Interface Tools", are the tools that provide functionalities for the toolbox GUI such as displaying decision machines, presenting cross-validation and test results, getting the value of tunable parameters from the user, checking the user-assigned values to the parameters, etc.
- **01_DecisionMachines**: This folder contains various functions for training and testing seven decision machines models: decision tree, support vector machine (SVM), random forest, ensemble $k$-nearest neighbors ($k$-NN), linear discriminant analysis (LDA), naïve Bayes, and neural network. The MATLAB machine learning toolbox is employed by these functions to provide simple and ready-to-use functionalities for file fragment classifications. In other words, the core of the learning algorithms is taken from MATLAB machine learning toolbox, and for user convenience, the parameters are set within the GUI of Fragments-Expert.
- **02_FeatureExtraction**: This folder contains the functions for extracting the features described in Section 2. As mentioned before, for the sake of run-time speed, some of these functions are written in C-MEX format. These features are categorized into nine subcategories that are located in nine subfolders as follows.

- **Lower-Order Statistics**: Basic lower-order statistics (described in Section 2-6), window-based statistics (described in Section 2-9), and autocorrelation (described in Section 2-10) are in this category.

- **Higher-Order Statistics**: These features are described in Section 2-7.

- **Frequency Domain Features**: These features are described in Sections 2-11 and 2-8.

- **Bit Patterns**: Audio patterns and *n*-grams are the feature types in this category. These features are described respectively in Sections 2-15 and 2-4.

- **Byte Distribution Features**: BFD, RoC, longest contiguous streak of repeating bytes, and byte concentration features are the feature types in this category. These features are described in Sections 2-1, 2-2, 2-3, and 2-5.

- **Byte Patterns**: Video patterns presented in Section 2-14 are in this category.

- **Randomness Features**: Binary ratio, entropy, false nearest neighbors, Lyapunov exponents, and Kolmogorov complexity are the feature types in this category. These features are described in Sections 2-12, 2-13, 2-17, and 2-16.

- **Similarity Features**: Longest common substrings, longest common subsequences, and centroid models are the feature types in this category. These features are described in Sections 2-19 and 2-20.

- **Texture Features**: These features are described in Section 2-18.

- **11_ScriptsForDecisionMachines**: This subfolder contains MATLAB script files for training, testing, and cross-validation of decision machines. These scripts are called from the GUI callback functions. Moreover, the scripts for loading decision machines and also the test and cross-validation results of decision machines are placed in this folder.

- **12_ScriptsforDatasets**: This subfolder contains MATLAB script files for generating and dealing with datasets, which includes generating fragments from raw multimedia files, generating a dataset of features, generating a dataset of features for a previously trained decision machine, and loading previously generated datasets. Also, the scripts for expanding, random permutation, and partitioning datasets are placed in this subfolder. Moreover, the scripts for merging labels in a dataset or extracting a sub-dataset from a dataset can be located in this subfolder. All these scripts are called from the GUI callback functions.

- **13_ScriptsforVisualization**: This subfolder contains two MATLAB script files. These scripts are called from the GUI callback functions. The details of these scripts are presented in Section 4-9.

  - Script `Script_Plot_FeatureHistogram_FFC` plots the histogram of feature values for one or more classes of samples.

  - Script `Script_Plot_Samples_in_FeatureSpace_FFC` displays the distribution of data samples in 2-Dimensional (2-D) or 3-Dimensional (3-D) feature spaces.

- **14_FeatureSelection**: This subfolder contains two MATLAB script files. These scripts are also called from the GUI callback functions. The details of these scripts are given in Section 4-4.

- Script Script_SequentialForward_FeatureSelection_FFC implements a sequential forward feature selection strategy.
- Script Script_SequentialForward_FeatureSelection_FFC implements an embedded feature selection strategy using decision tree model.

# 4  Toolbox Functionalities

By running `Main_FFC`, the GUI of Fragments-Expert will open. The overall appearance of the GUI is shown in Figure 3. In the following sub-sections, we present the toolbox functionalities. All these functionalities are available from the designed GUI.

Dataset generation and manipulation is the main capability of Fragments-Expert. You can generate a dataset of fragments from raw multimedia files in the form of generic binary data files. After that, these binary files can be used to generate a dataset of file fragment features. Many feature types, including all commonly used features, are available in the toolbox. You can also generate a dataset for an already trained decision machine.

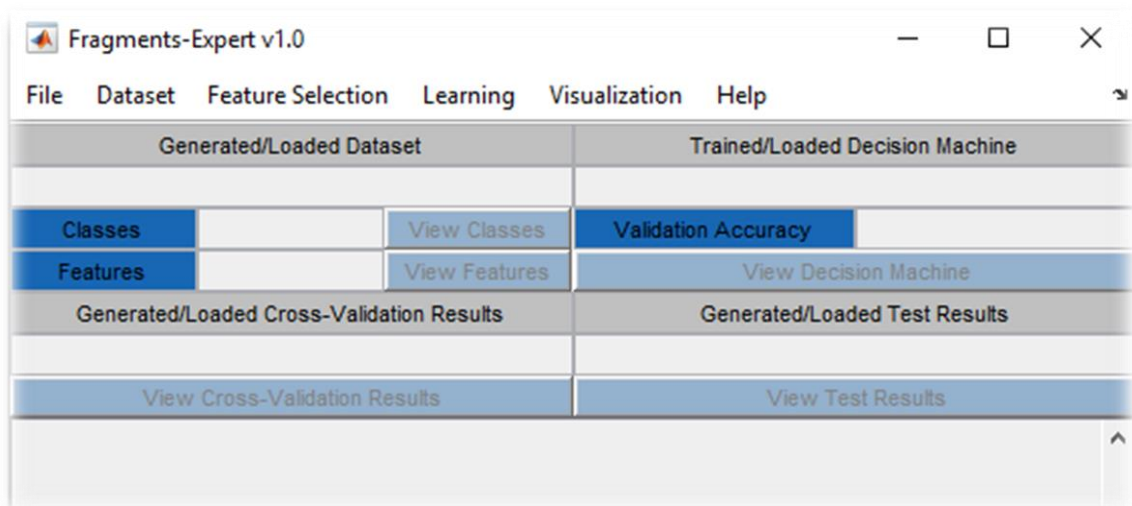

Figure 3: The overall appearance of Fragments-Expert GUI.

## 4-1    Generating Dataset of File Fragments

To extract fragments from raw multimedia files, click on the "Convert Raw Multimedia to Fragments Dataset" submenu in the "Dataset" menu. By clicking on this item, a new window, as shown in Figure 4, will open. In this stage, you should choose the parameters for generating the fragments as follows.

- The first input is an integer (or a vector of integers) representing the size(s) of generated fragments in bytes. If you specify a vector, the generated fragments are with lengths that are taken randomly and independently from this vector.
- The second input determines the percent of fragments from the beginning of the file that should be discarded

- The third input determines the percent of fragments from the end of the file that should be discarded
- The last input is the maximum number of fragments taken from each file.

Figure 4: Parameters for generating the fragments.

After filling the parameters in and hitting the OK button, you are prompted to select the main folder that contains the multimedia files. Note that this should be the directory that contains some folders, in which the raw files are stored. It is assumed that the file types in each subfolder are the same. After that, you must select the folder in which you want the fragments to be stored. Note that the second folder that you select should not be a subfolder of the first folder. In other words, you must choose a completely different directory or the action will be aborted. After that, the file fragments generation starts. Corresponding to each subfolder, a generic binary data file with .dat extension is created that contains the fragments taken from the files of that subfolder. The total number of fragments for each class (i.e. each subfolder) is shown in the main text window of the toolbox.

### 4-2    Generating Dataset of Features

For generating a dataset of features from generic binary data files, click on the "Generate Dataset from Generic Binary Files of Fragments" submenu in the "Dataset" menu. In the next window, a list of feature types is shown as follows.

- Byte Frequency Distribution
- Rate of Change
- Longest Contiguous Streak of Repeating Bytes
- $n$–grams
- Byte Concentration Features: Low, ASCII, and High
- Basic Lower-Order Statistics: Mean, STD, Mode, Median, and MAD
- Higher-Order Statistics: Kurtosis and Skewness
- Bicoherence
- Window-Based Statistics

- Auto-Correlation
- Frequency Domain Statistics (Mean, STD, Skewness)
- Binary Ratio
- Entropy
- Video Patterns
- Audio Patterns
- Kolmogorov Complexity
- False Nearest Neighbors
- Lyapunov Exponents
- GIST Features
- Longest Common Subsequence
- Longest Common Substring
- Centroid Model

From the displayed list, select the features that you want to be extracted from your fragments. You can multi-select from the list using the Ctrl key. After that, hit the OK button. If your selected features require any parameters, another window will be opened to fill in the parameters. Then you must select the generic binary data files containing the fragments. Each file in .dat format represents a category (i.e. a class label). A progress bar is shown during reading the fragments. After reading the fragments, the toolbox wants you to confirm variable names corresponding to class labels. Here you can choose any desired name for each category. After confirming the names, the feature extraction process starts. Once, this process is completed, you should save the generated dataset of features. The dataset will be loaded in the toolbox environment under the "Generated/Loaded Dataset" section (see Figure 5); from there, you can view the classes and features. The number of data samples in each class is equal to the number of fragments in the corresponding .dat file.

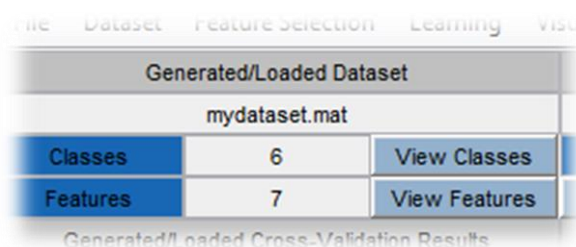

**Figure 5: An example of a generated dataset with six classes and seven features in the toolbox environment.**

## 4-3 Generating Dataset for a Decision Machine

You may have a previously trained decision machine and now you want to generate a new dataset that its feature set is matched to this already trained decision machine. To do that, first, the previously trained machine must be loaded in the toolbox (see Section 4-10). Then click on the "Generate Dataset (for Decision Machine) from Generic Binary Files of Fragments" submenu in the "Dataset" menu.

The next steps are quite similar to the steps explained in section 4-2; except now you don't need to select any feature type. The dataset is generated automatically and then, after saving on the disk, it is loaded in the toolbox environment.

## 4-4  Feature Selection

When you create a dataset of features, you may have many features that most of them might be useless in the classification process. To obtain more relevant features and achieve less computational cost in training decision machines, we need to employ some feature selection methods. Sometimes as an expert in the field, you can determine the features that contribute most to your classification scenario. However, manually selecting the features could be very hard or time-consuming. In Fragments-Expert, we have implemented two useful and simple feature extraction methods.

### *Embedded Method*

Embedded methods refer to techniques that select the features during the training phase. The learning algorithm itself selects the features as a step of the learning. The most typical embedded method is the decision tree.

In each training step, a decision tree selects the best feature based on a goodness measure. So, as we go upper in a trained tree, the decisions are made based on more relevant features. We can take advantage of this property and turn the decision tree into a feature selector.

To use this method, first, the dataset must be loaded in the toolbox (see Section 4-10). Then click on the "Embedded: Decision Tree" submenu in the "Feature Selection" menu. Since you are about to train a decision tree, you must input the parameters similar to the parameters described in Section 4-6. After the parameters are set and the training phase is completed, you will see a list of selected features sorted based on the relative node sizes. You can select among these features. Finally, the new feature set is created and loaded in the toolbox after saving it on the disk.

### *Wrapper Method*

Wrapper methods search through different possible subsets of the feature set. These methods evaluate each subset by the accuracy of the learning algorithm. Note that for large problems, this method is very time-consuming.

In Fragments-Expert we have employed LDA as the core learning algorithm for this type of feature selection method. To apply this method, first, load the dataset in the toolbox environment. Then, click on the "Wrapper: Sequential Forward Selection with LDA" submenu in the "Feature Selection" menu. LDA is used with cross-validation in a forward selection scenario to select the features. To get an understanding of LDA and cross-validation refer to Sections 4-6 and 4-8. During the feature selection process, the included features and corresponding accuracies are listed in the toolbox. After the process is completed, you can see a list of selected features. You can select the final selected features among these features.

## 4-5  Operations on Datasets

In the Fragments-Expert environment, you can apply some operations on a dataset including random permutation of samples in a dataset, merging two datasets, merging class labels within a dataset, and generating sub-dataset from a dataset.

### *Random Permutation of a Dataset*

This process randomly permutes the samples in a dataset. Each sample contains the features that describe a specific fragment. Note that the permutation is performed in a way that the fragments of a single file stay together.

To permute the samples in a dataset, first, the dataset must be loaded in the toolbox. Then, you should click on the "Random Permutation of Dataset" submenu in the "Dataset" menu. Then you must save the permuted dataset.

### *Merge Two Datasets*

This part of Fragments-Expert helps you to add features to an already created dataset. In this case, the dataset with the new feature set of the same fragments must be loaded in the toolbox. To add the features of this loaded dataset to an already saved dataset, click on the "Expand Dataset" submenu in the "Dataset" menu. Then you must select the old dataset. Note that both datasets must have matching dataset sizes, output labels, and file identifiers.

### *Merge Labels in a Dataset*

Sometimes you may want to merge labels in a dataset to form more general class labels. To do so, the dataset must be loaded first. Then click on the "Merge Labels in Dataset" submenu in the "Dataset" menu. A window will open. There you can multi-select the classes you want to merge their labels. If this process needs to be done for other classes click on the OK button, if not click on the cancel button. In Figure 6, an example is shown. In this example, we are merging the samples of AAC codec with different bitrates into one general label AAC. After selecting the merged labels, you need to confirm the labels for the merged classes. In Figure 7, an example is shown.

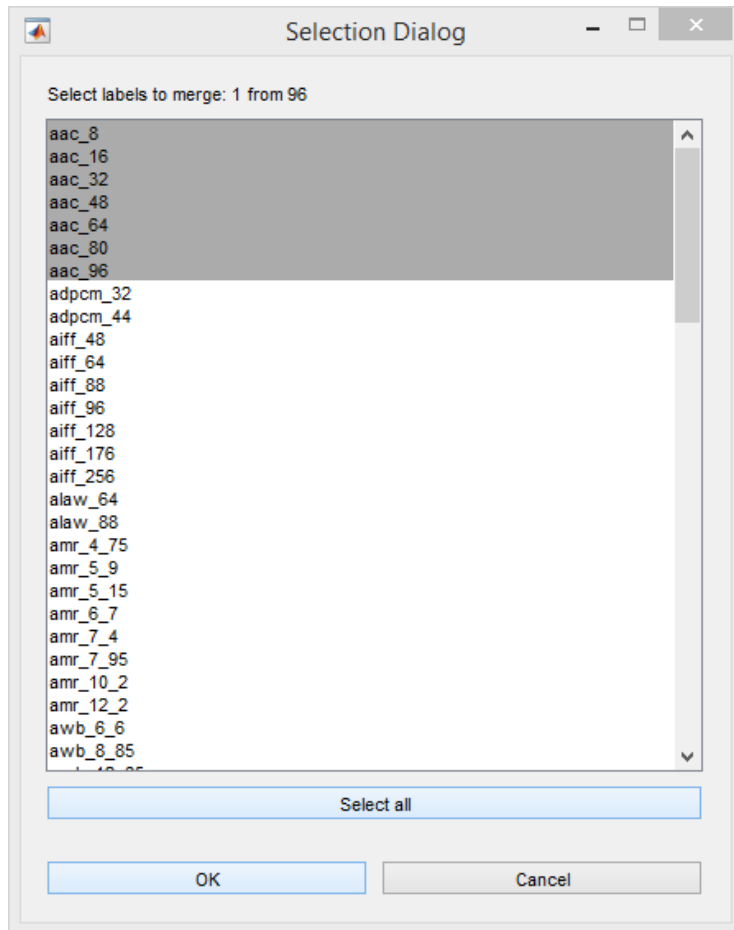

Figure 6: An example of selecting seven different labels to merge them into one class label.

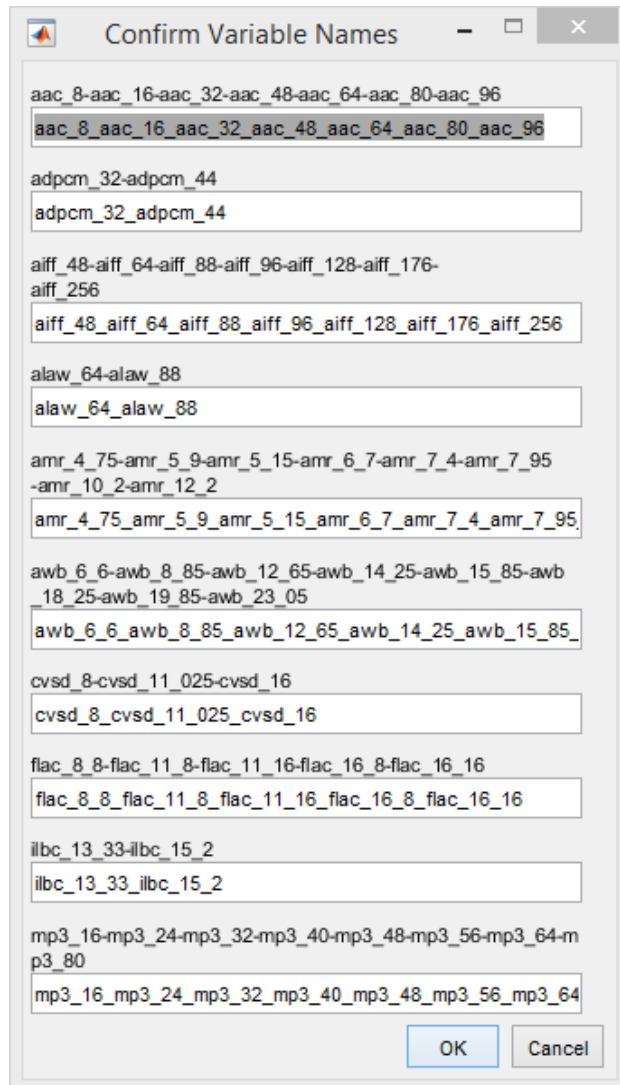

Figure 7: An example of confirming labels for merged class labels.

### *Generating Sub-Datasets from a Dataset*

This feature of Fragments-Expert helps you to select a subset of dataset classes and/or features. When the original dataset is loaded, click on the "Select Sub-Dataset" submenu in the "Dataset" menu. In the next window, you can multi-select the classes you want to keep. Then you can select among features. For both classes and features, if you want to keep all of them, select all.

## 4-6 Train a Decision Machine for File Fragments Classification

Using Fragments-Expert, you can apply several machine learning algorithms and methods on your dataset. You can train, test, and cross-validate a model. The currently available machines are decision tree, SVM, random forest, ensemble *k*-NN, linear discriminant analysis, naïve Bayes, and neural network. In the following sections, more details are presented.

To train a decision model, first, your dataset must be loaded in the toolbox environment. Then you should click on the "Train Decision Machine" submenu in the "Learning" menu. In the next window, you should select a decision model among decision machines. After selecting the decision model, you must provide some parameters. For each model, some default parameters are set; however, you may want to modify them to what works best for your experiment.

Three general parameters are required for all decision machines:

- The weighting method: This parameter determines whether the frequency of instances of each class label should be considered or not. If the weighting method "balanced" is chosen, the sample weights are set in a way that implies similar importance of all classes in the learning process. On the other hand, if the weighting method "uniform" is chosen, the classes with a larger number of samples are considered as more important classes.

- Start and end of the train/validation in the dataset: This parameter is a vector of length two with elements in the range of [0 1] that determines the relative position of the start and the end of the train/validation in the dataset. Note that 0 corresponds to the first sample in the dataset and 1 corresponds to the last sample in the dataset.

- Train and validation percentages taken from the dataset: This parameter is a vector of length two with a sum of elements equal to 100. The first element determines the percent of training data in the train/validation set. According to the toolbox settings, at least 70% of the train/validation set should be dedicated to the training phase.

For most of the decision models, the method for feature scaling is also prompted. The scaling method can be either standardization (also called z-score) or min-max normalization. Assume that we have $S$ samples in the training phase, where each of them is defined by $F$ features. For example, assume that $f_{i,j}; j = 1,2, \dots, S$ are the values of the features $f_i; i = 1,2, \dots, F$ over all samples. If we denote the scaled feature values by $\hat{f}_{i,j}$, for z-score scaling, we have

$$\hat{f}_{i,j} = \frac{f_{i,j} - \mu_{f,i}}{\sigma_{f,i}}, \tag{30}$$

where $\mu_{f,i}$ and $\sigma_{f,i}$ are respectively the mean and standard deviation for values $f_{i,j}; j = 1,2, \dots, S$. Furthermore, for min-max normalization we have.

$$\hat{f}_{i,j} = \frac{f_{i,j} - a_{f,i}}{b_{f,i} - a_{f,i}}, \tag{31}$$

where $a_{f,i}$ and $b_{f,i}$ are respectively the minimum and maximum values among $f_{i,j}; j = 1,2, \dots, S$.

After setting all parameters for training and pressing the OK button, a progress bar indicates the progression of the training process and the remaining time. After the training phase completion, the trained model is loaded in the toolbox environment. Also, the training parameters and results are shown in the main text window of the toolbox. The confusion matrices for training and validation are also shown in the command window. Figure 8 displays the results of an example with the LDA decision model.

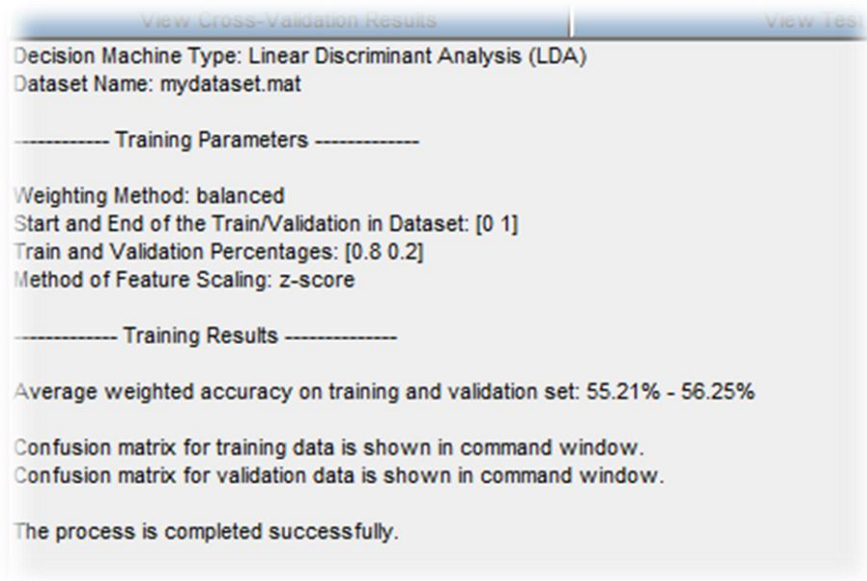

**Figure 8: An example displaying training parameters and results in the main text window of the toolbox.**

### *Train a Decision Tree*

To train a decision tree, besides the general parameters, you need to input the minimum relative number of leaf node observations to total samples. This relative number is then multiplied by the size of the training set. The result is the minimum number of observations per tree leaf. Note that the validation percent for the decision tree should be at least 15%.

When the training is completed, a figure window will open to display the trained tree. Also, you are prompted to save the trained model. In Figure 9 an example is shown.

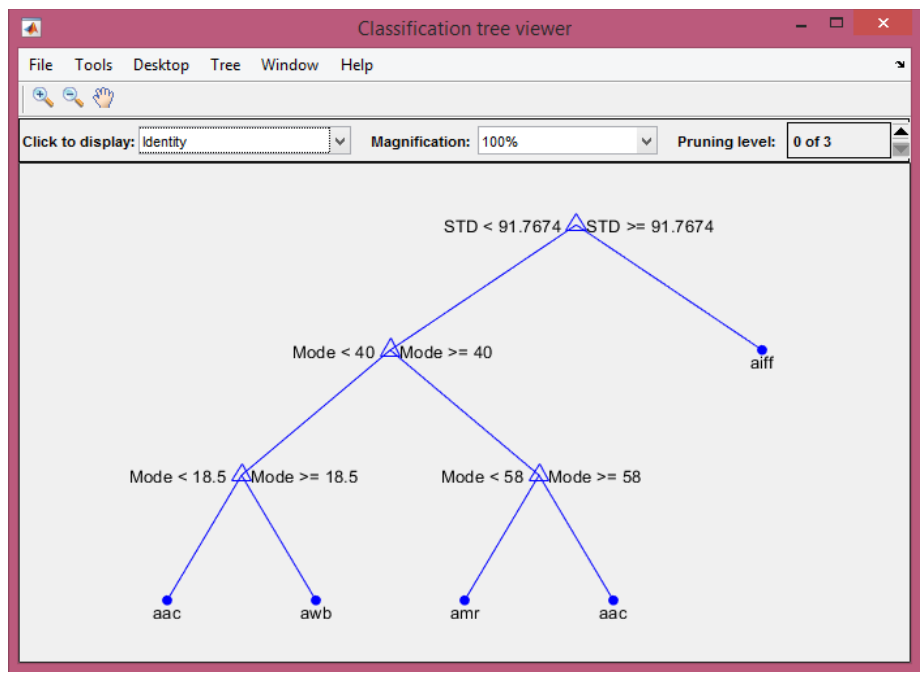

**Figure 9: A very simple trained decision tree.**

### Train a Multi-Class Support Vector Machine Classifier

In Fragments-Expert, a multi-class SVM classifier is trained by a one-versus-all strategy, in which one binary classifier is trained per class. To train a multi-class SVM, besides the general parameters, you need to provide some other parameters. These parameters are as follows.

- Value for box constraint in SVM: This number should be a positive real number.
- Kernel function for SVM: You can choose rbf, linear, or polynomial.
- Polynomial order for polynomial kernel function: This number, which is denoted by $q$, should be an integer from 1 to 7.
- Value for scaling kernel of SVM: This number, which is denoted by $c$, should be a positive real number.

SVM algorithm assigns a box constraint to each observation in the training data. A box constraint is a parameter that controls the maximum penalty imposed on margin-violating observations. A kernel function is used to compute the elements of the Gram matrix. Each element of the Gram matrix is an inner product of the transformed predictors (i.e. features) using the kernel function. Suppose $G(\hat{\mathbf{f}}_j, \hat{\mathbf{f}}_k)$ is element $(j, k)$ of the Gram matrix, where $\hat{\mathbf{f}}_j = [\hat{f}_{1,j}, \hat{f}_{2,j}, \ldots, \hat{f}_{F,j}]$ and $\hat{\mathbf{f}}_k = [\hat{f}_{1,k}, \hat{f}_{2,k}, \ldots, \hat{f}_{F,k}]$ are $F$-dimensional normalized feature vectors representing samples $j$ and $k$ in the training set. In Table 4, a brief description of kernel functions is given.

**Table 4: Different SVM kernels.**

| Kernel Function | Description | Formula |
| --- | --- | --- |
| rbf | Radial Basis Function | $G(f_j, f_k) = e^{-\|\hat{\mathbf{f}}_j - \hat{\mathbf{f}}_k\|^2 / c^2}$ |
| linear | Linear | $G(f_j, f_k) = \hat{\mathbf{f}}_j \hat{\mathbf{f}}_k^T / c^2$ |
| polynomial | A polynomial with order $q$ | $G(f_j, f_k) = (1 + \hat{\mathbf{f}}_j \hat{\mathbf{f}}_k^T / c^2)^q$ |

### Train a Random Forest

To train a Random Forest, besides the general parameters, you need to choose the number of trees in the random forest and the minimum relative number of leaf node observations to total samples in each tree. This relative number is then multiplied by the size of the training set. The result is the minimum number of observations per tree leaf.

### Train an Ensemble of k-Nearest Neighbors Classifiers

To train an ensemble *k*-NN classifiers, besides the general parameters, you need to provide some other parameters. These parameters are as follows:

- the number of randomly selected features for each *k*-NN learner,
- the number of *k*-NN learners in the ensemble,
- and the number of nearest neighbors for classifying each sample.

### *Train a Naïve Bayes Classifier*

For this classifier, no additional input parameters are asked by the toolbox. The naïve Bayes model is trained with the following default parameters.

- Data distribution: Kernel smoothing density estimate is used to model the data.
- Kernel smoother type: Gaussian is set as the kernel smoother type.

### *Train a Linear Discriminant Analysis Classifier*

You can use the LDA classifier to classify fragments based on their feature distribution. The model assumes data has a Gaussian mixture distribution. Since the discriminator in the current version of Fragments-Expert is pseudo-linear, the model assumes the same covariance matrix for each class, and only the means vary.

### *Train a Neural Network*

You can train a neural network to classify training data. In the current version of Fragments-Expert, two-layer pattern recognition neural network model is considered. So, besides the general parameters, you need to input the dimension of the hidden layer.

The training function updates weight and bias values according to the scaled conjugate gradient method. Moreover, cross-entropy is used as a measure for network performance.

## 4-7    Test a Trained Decision Machine

To test any trained model, the already trained machine and a compatible dataset must be loaded in the toolbox environment. Note that the dataset must be compatible; i.e. the feature set of the dataset must be the same as the feature set used for training the model. You must provide the start and the end of the test samples in the dataset and also the weighting method.

After the test procedure is completed, the test result can be saved. The test parameters and results are shown in the main text window of the toolbox.

Figure 10 shows an example. In this example, we have trained a decision tree for an audio codec dataset. The dataset contains 20 audio file formats and the first 256 BFD features along with the first 256 RoC features are considered as features. Now we want to test the performance of this trained tree for classifying image file formats. We generate a dataset of features for four image file formats using the procedure explained in Section 4-3. After running the test procedure, the test result is displayed. Moreover, you can save these results.

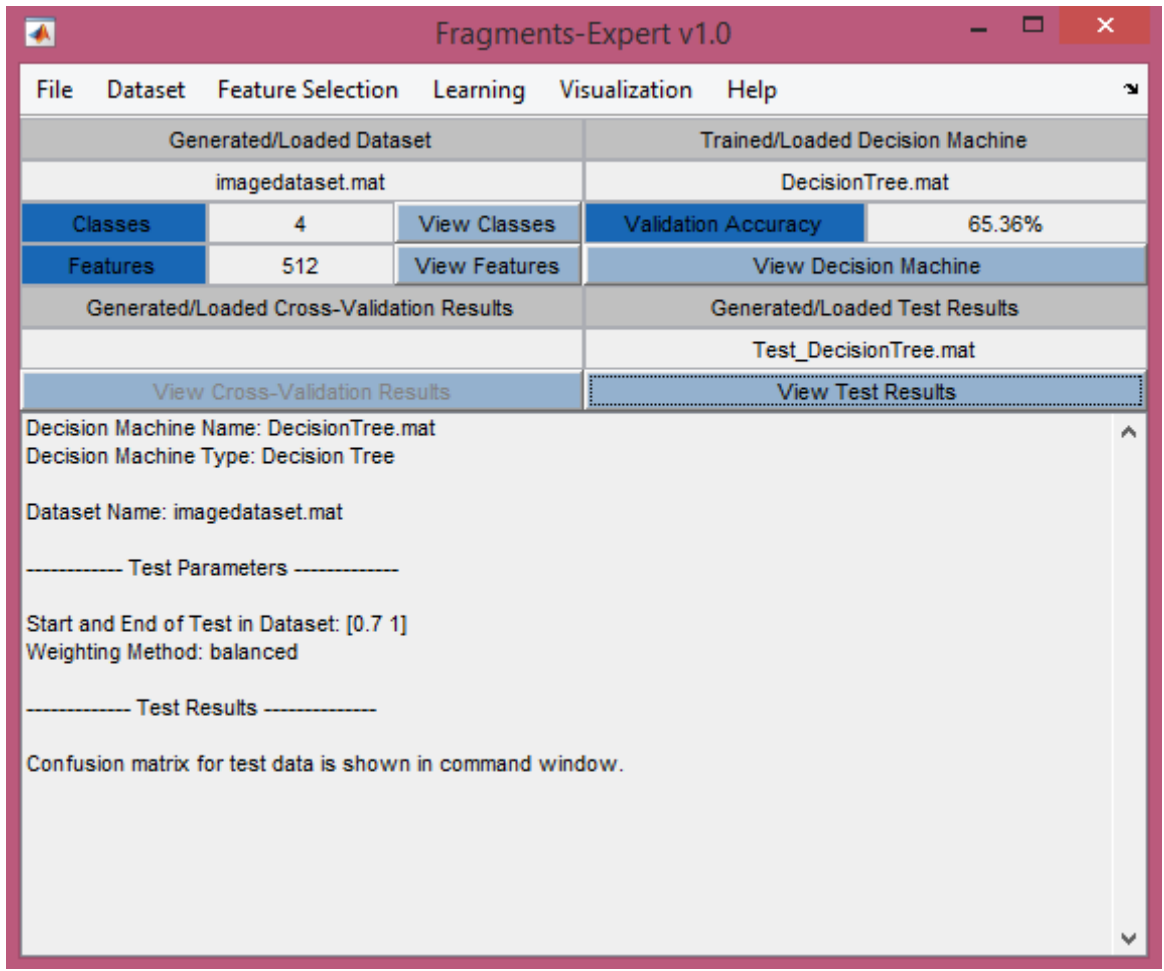

Figure 10: The toolbox environment after the test procedure.

## 4-8 Cross-Validation for a Decision Model

To obtain the optimal parameters for a decision model or to assess the average performance of a learning method, you can use cross-validation. To use cross-validation, first, your dataset must be loaded in the toolbox environment. Then, you should click on the "Cross-Validation of Decision Machine" submenu in the "Learning" menu. In the next window, you can select among decision machines. Besides the parameters for each machine, you must choose the K value for K-fold cross-validation. When the procedure is completed, the cross-validation result is shown in the main text window of the toolbox. The cross-validation confusion matrices are also shown in the command window of MATLAB.

## 4-9 Data Visualization

Visualization tools are available to give a better understanding of the distribution of feature values among data samples. In the current version of Fragments-Expert, you can plot a feature histogram or display the distribution of data samples in feature space.

### *Plot Feature Histogram*

You can plot the histogram of one or more features for one or more selected class labels. To do so, click on the "Plot Feature Histogram" submenu in the "Visualization" menu. In the next window, select among features. For each chosen feature a separate histogram will be plotted. After confirming the feature label(s), you must choose the desired class label. You can select more than one class. Now you must again confirm the class labels. In the next window, you should set the number of bins for the histogram and the subplot organization. After that, press the OK button. Figure 11 shows an example of a histogram.

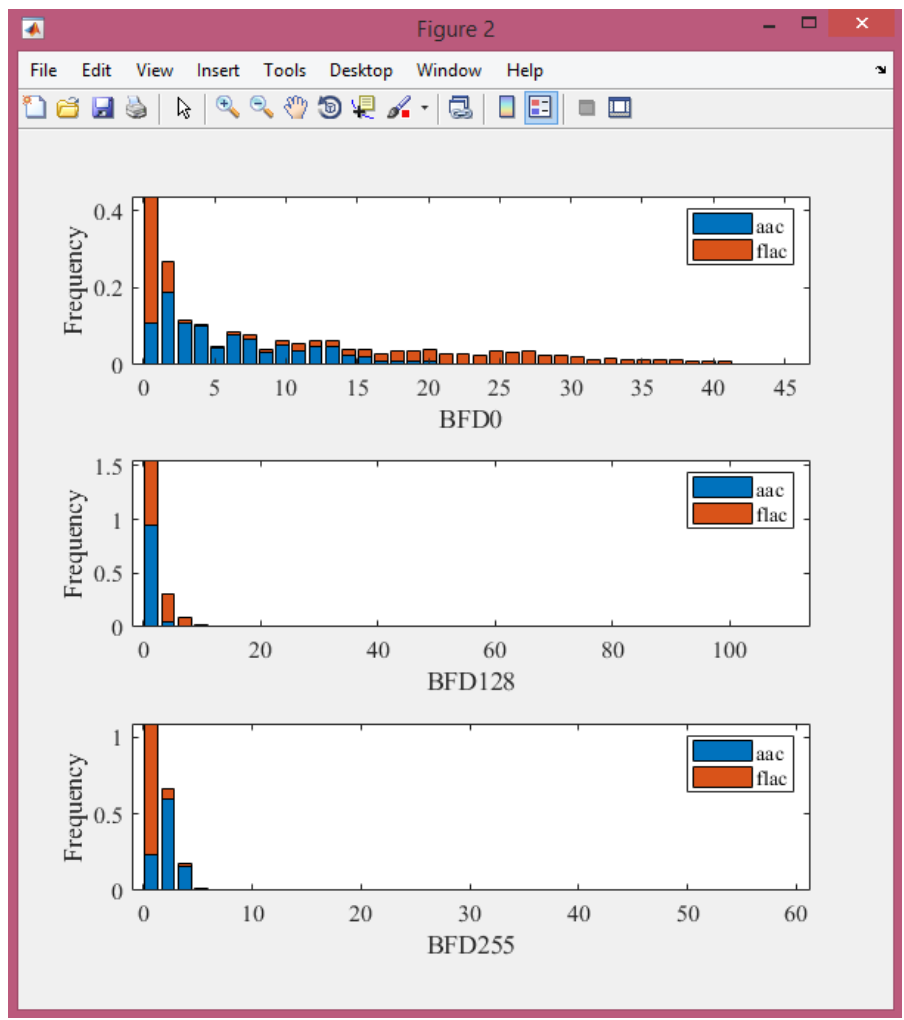

Figure 11: An example of a histogram plot.

### *Display Samples in Feature Space*

You can display samples of multiple class labels in a feature space. You can select two or three features to form a 2-D or 3-D feature space, in which, each sample of the dataset is represented as a data point.

To do so, click on the "Display Samples in Feature Space" submenu in the "Visualization" menu. In the next window select two or three features. Also, you should select some classes. In this process, you must also confirm the feature and class labels. Afterward, you can see the distribution of samples in the selected feature space. In Figure 12, an example of a 3-D feature space representation is shown.

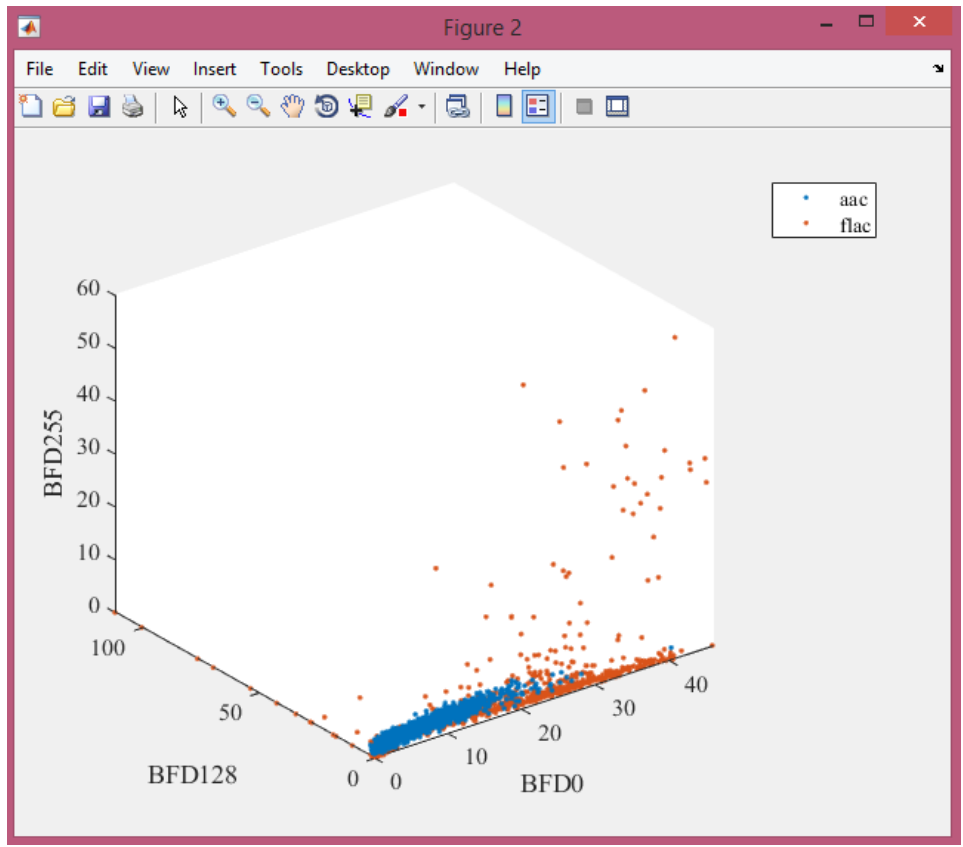

**Figure 12: An example of displaying samples in a 3-D feature space.**

## 4-10  Loading Previously Generated Data and Results

If you have saved a dataset, a decision machine, a test result, or a cross-validation result, you can load them again in the toolbox.

### *Load Dataset*

To load a dataset, click on the "Load Dataset" submenu in the "File" menu. Select the dataset in your device. When the dataset is loaded, you can view class labels and features.

### *Load Decision Machine*

To load a decision machine, click on the "Load Decision Machine" submenu in the "File" menu. Select the trained machine. When the decision machine is loaded, you can see the validation accuracy.

### *Load Test Results*

To load a test result, click on the "Load Test Results" submenu in the "File" menu. Select the test results. When it is loaded, you can click on "View Test Results" to see the test results.

### *Load Cross-Validation Results*

To load a cross-validation result, click on the "Load Cross-Validation Result" submenu in the "File" menu. Select the cross-validation results. When it is loaded, you can click on "View Cross-Validation Results" to see the result of cross-validation. In this case, you can see the confusion matrices in the command window.

# 5  An Illustrative Example

In this section, we present an example of file fragment classification for textual file formats. We employ the dataset presented in [43]. We consider five file formats DOC, DOCX, PDF, RTF, and TXT. For simplicity, we consider only the fragments of the English textual files. So, the data files corresponding to these five formats are DOC-EN.dat, DOCX-EN.dat, PDF-EN.dat, RTF-EN.dat, and TXT-EN.dat. Each data file contains 1500 fragments. The length of all fragments is equal to 1024 bytes.

## 5-1    Generating Dataset of Features

For generating a dataset of features from the above-mentioned generic binary data files, we click on the "Generate Dataset from Generic Binary Files of Fragments" submenu in the "Dataset" menu. Then, as shown in Figure 13, a window is opened. In this window, a list of feature types is shown. We select the following features and press the OK button: BFD, RoC, the longest contiguous streak of repeating bytes, $n$-grams, byte concentration features, basic lower-order statistics, higher-order statistics, window-based statistics, auto-correlation, frequency domain statistics, and entropy.

Then, as shown in Figure 14, another window is opened to fill in the parameters for the four categories of features: $n$-grams, window-based statistics, frequency domain statistics, and auto-correlation. For $n$-gram, we choose to extract both 2-grams and 3-grams features. For window-based statistics, we choose a window size of 256 bytes. For frequency-domain statistics, the number of sub-bands is set equal to 4. Moreover, for the auto-correlation feature, the maximum lag value is set equal to 5. After, setting the parameters for feature extraction, we press the OK button.

After that, as shown in Figure 15, a window is opened to get the data files of fragments. In this stage, we select the following files and press the OK button: DOC-EN.dat, DOCX-EN.dat, PDF-EN.dat, RTF-EN.dat, and TXT-EN.dat. After reading the fragments, as shown in Figure 16, the toolbox shows a text box to confirm the variable names corresponding to five class labels. We confirm the default variable names DOC_EN, DOCX_EN, PDF_EN, RTF_EN, and TXT_EN.

After confirming the variable names, the feature extraction process starts. Once, this process is completed, the user is prompted to save the generated dataset (see Figure 17 ). We choose the default name "mydataset.mat" and press the save button. Now, the dataset is also loaded in the toolbox environment. Under the "Generated/Loaded Dataset" section, we can see that the dataset consists of five classes and 566 features. By pressing the "View Features" button, the list of the features is shown (see Figure 18).

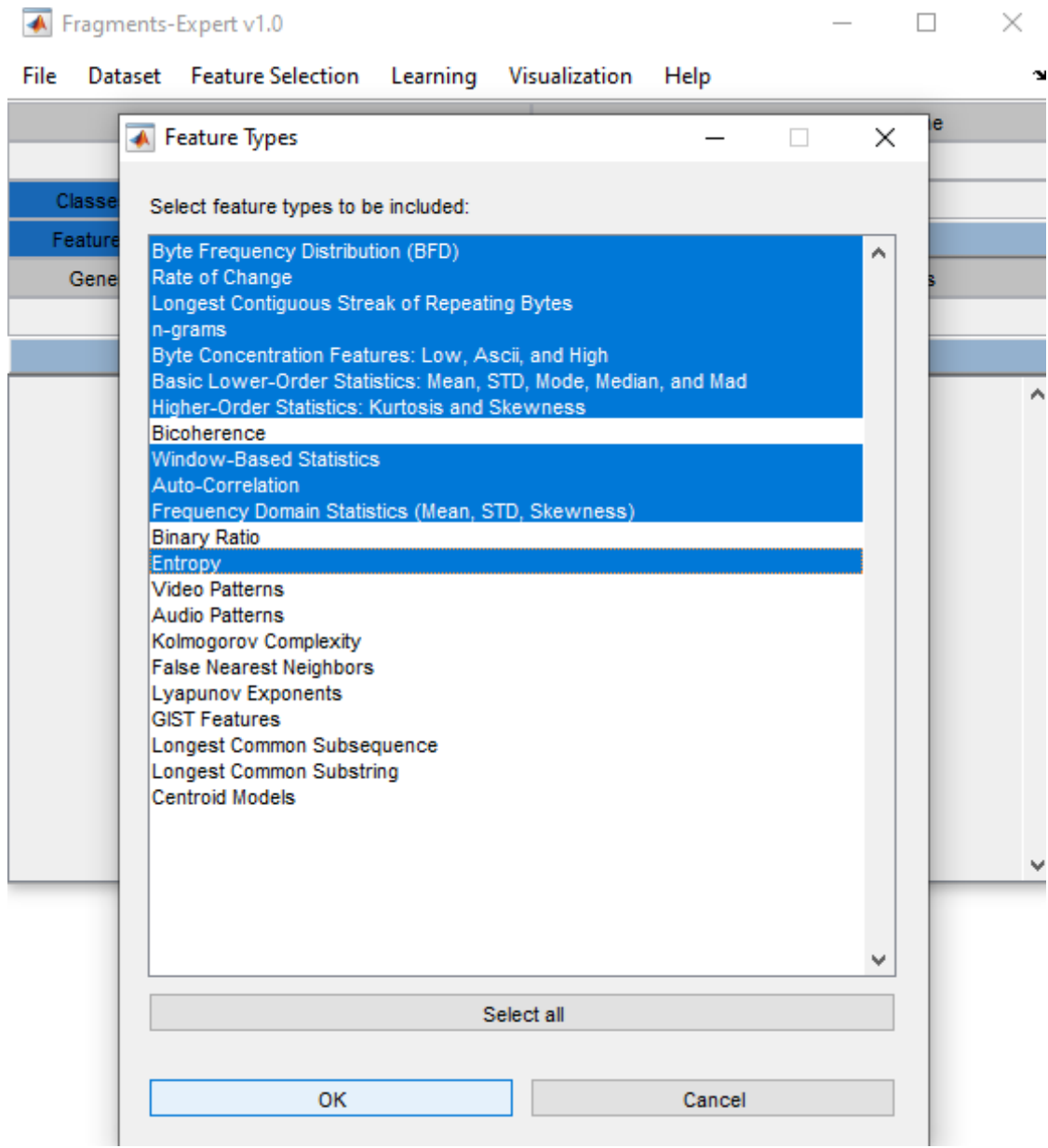

Figure 13: Illustrative example: selecting the features to be extracted.

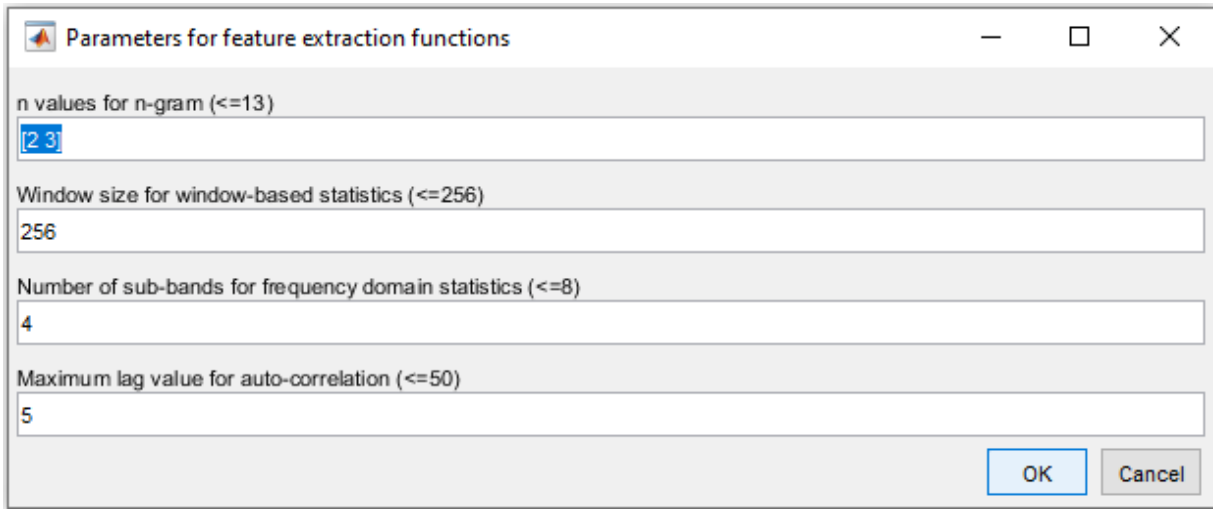

Figure 14: Illustrative example: determining the parameters for feature extraction.

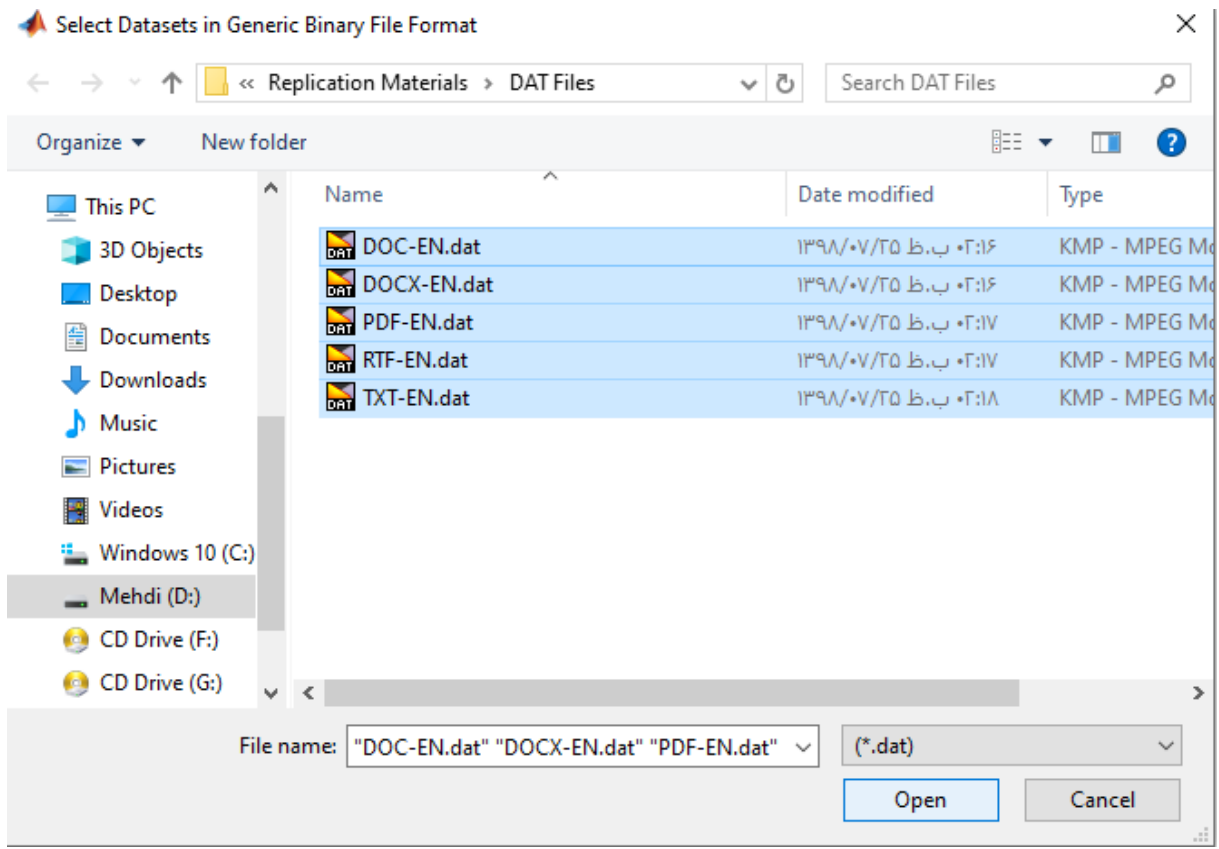

Figure 15: Illustrative example: selecting data files of fragments.

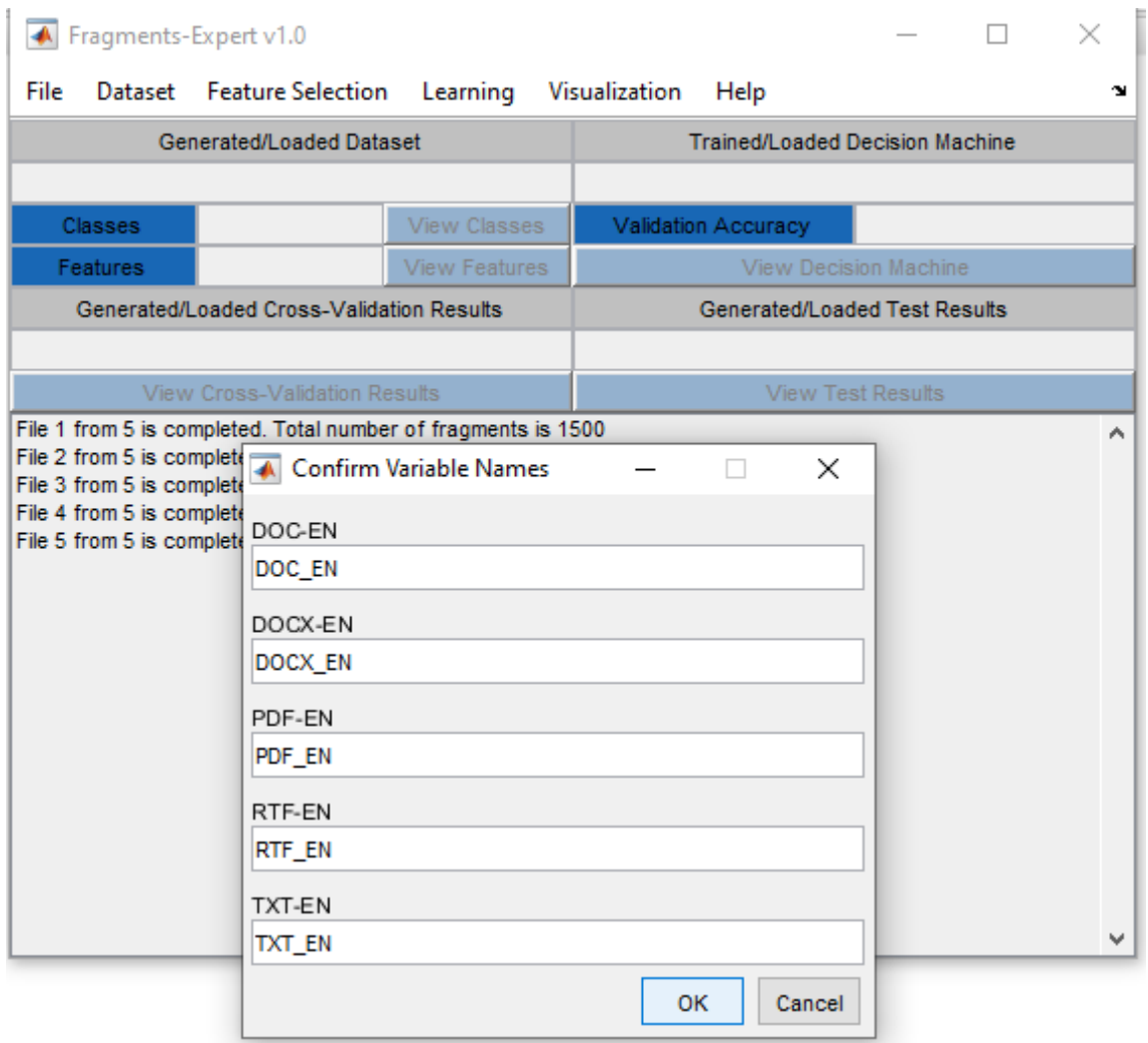

**Figure 16: Illustrative example: confirm variable names for class labels.**

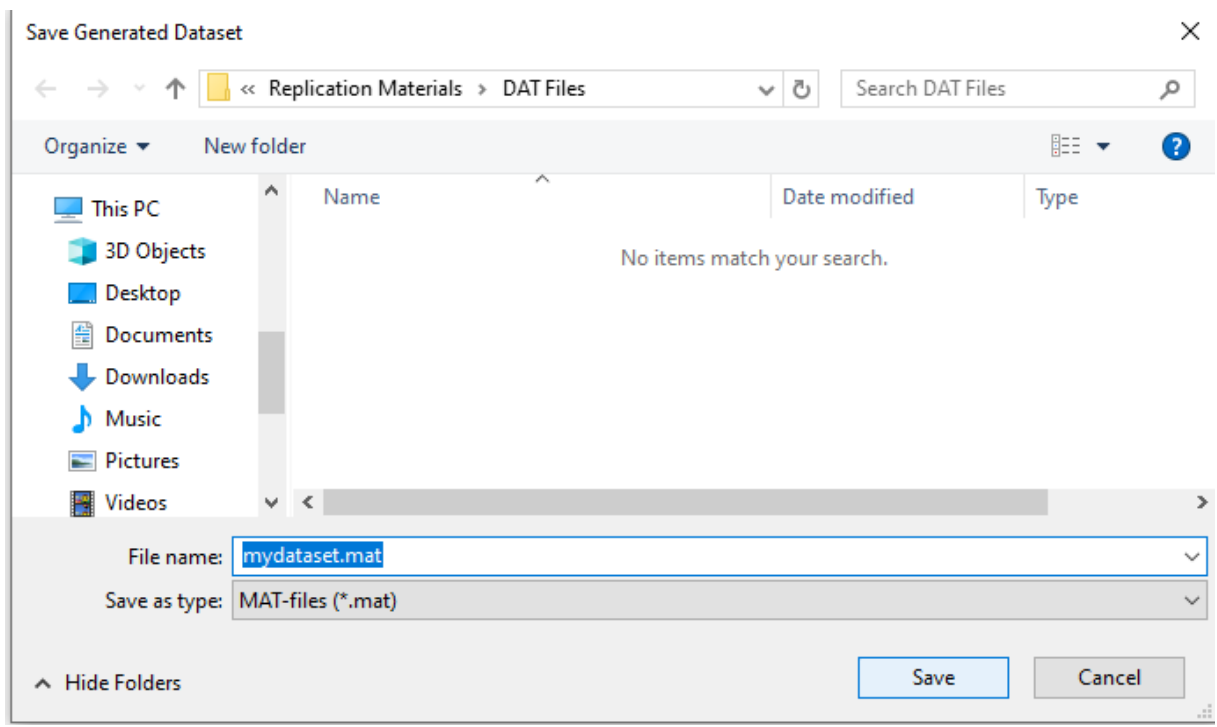

Figure 17: Illustrative example: saving the generated dataset of features.

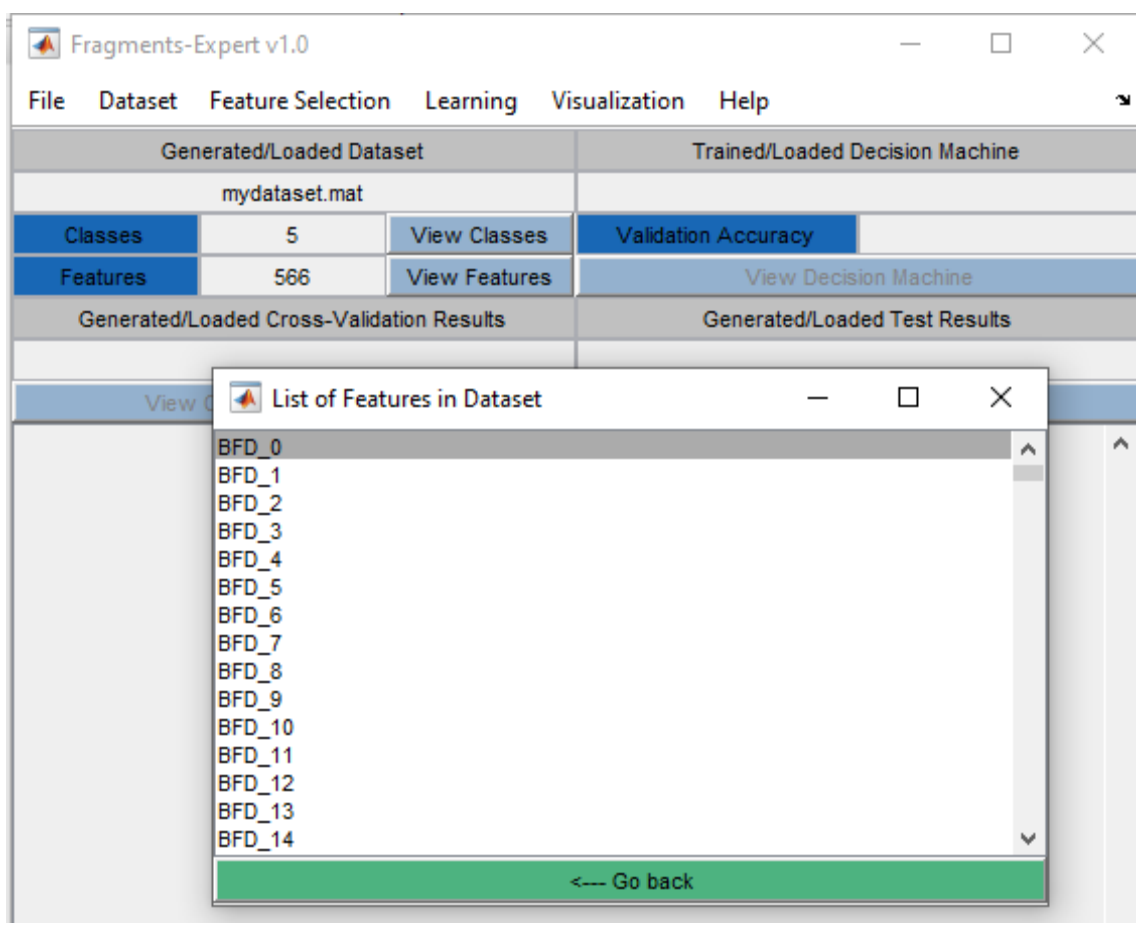

Figure 18: Illustrative example: viewing the list of features in the generated dataset.

## 5-2    Training a Decision Tree

For training a decision tree for our dataset, we click on the "Train Decision Machine" submenu in the "Learning" menu. As shown in Figure 19, a window is opened in which we select "Decision Tree" and press the OK button. As shown in Figure 20, we should set the training parameters in the next window. In this stage, the weighting method "balanced" is chosen. Also, the start and end of the train/validation in the dataset is set equal to [0 1], which indicates using all available samples for train/validation. Train and validation percentages are chosen to be equal to 80% and 20%, respectively. Finally, the minimum relative number of leaf node observations to total samples is set equal to 0.001. Since 80% of total samples (i.e. 6000 samples) are used for training, the minimum number of leaf node observations is considered to be equal to 6. To prevent overfitting, 1500 validation data samples (i.e. 20% of all samples) are used to prune the decision tree.

After setting the training parameters and pressing the OK button, a progress bar indicates the progression of the training process of the decision tree. After completion of the training phase, the trained decision tree is shown (see Figure 21). Moreover, the user is prompt to save the decision model (see Figure 22). After saving the decision model, the decision model is loaded in the toolbox environment. Also, the training parameters and results are shown in the main text window of the toolbox. As shown in Figure 23, the average accuracies for training and validation sets are around 93% and 85%, respectively. The confusion matrices for training and validation are also shown in the command window. As can be seen in Figure 23, the highest accuracies are obtained for DOC and RTF format.

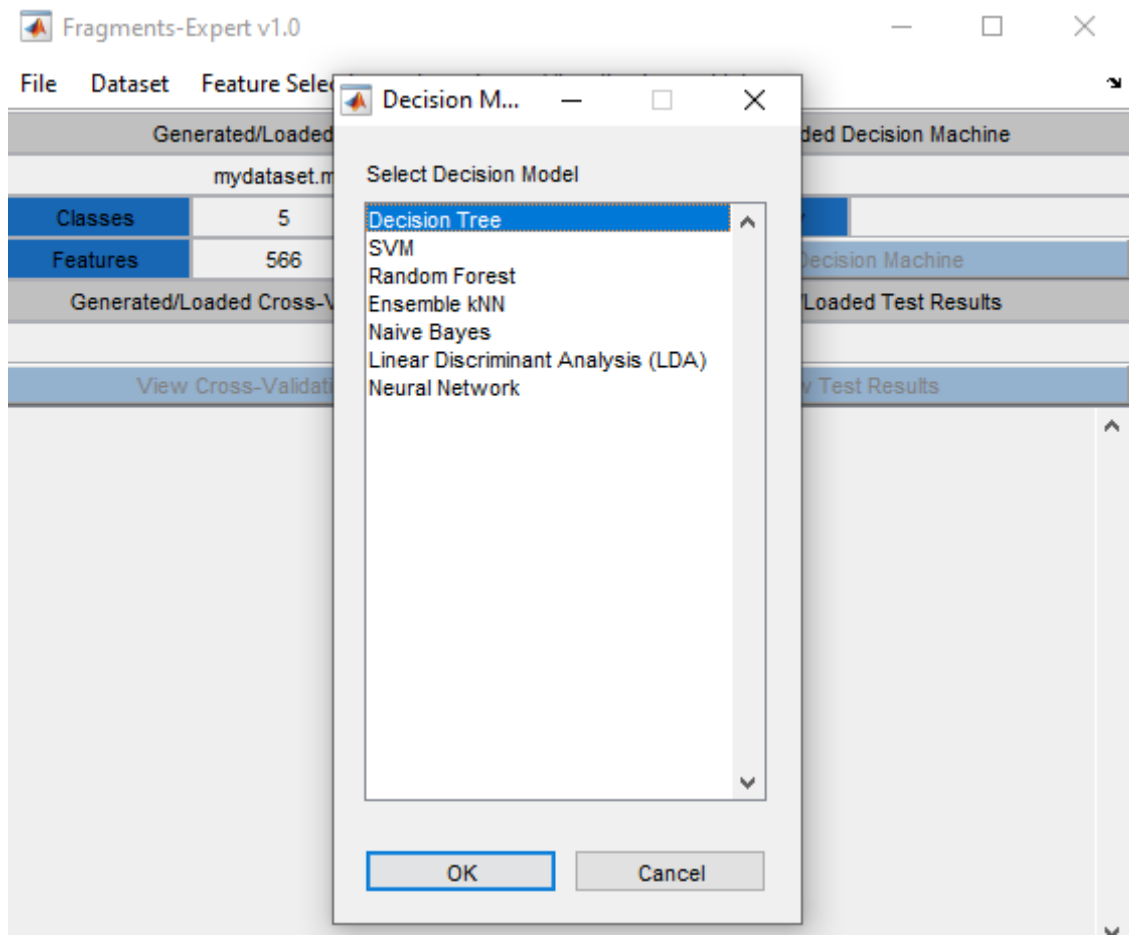

Figure 19: Illustrative example: selecting the decision model.

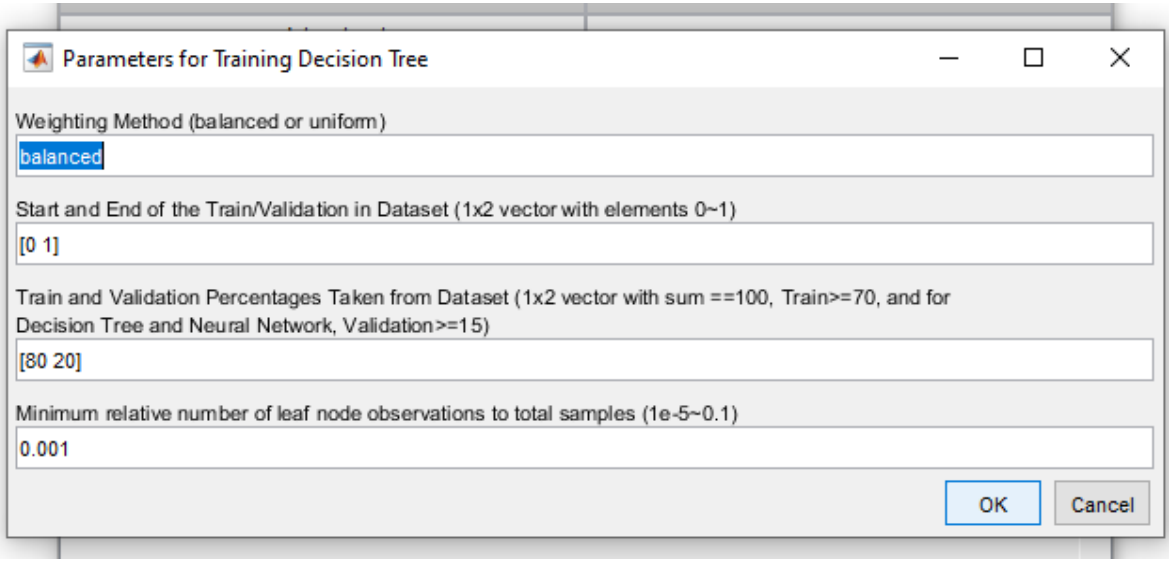

Figure 20: Illustrative example: setting the parameters for training the decision model.

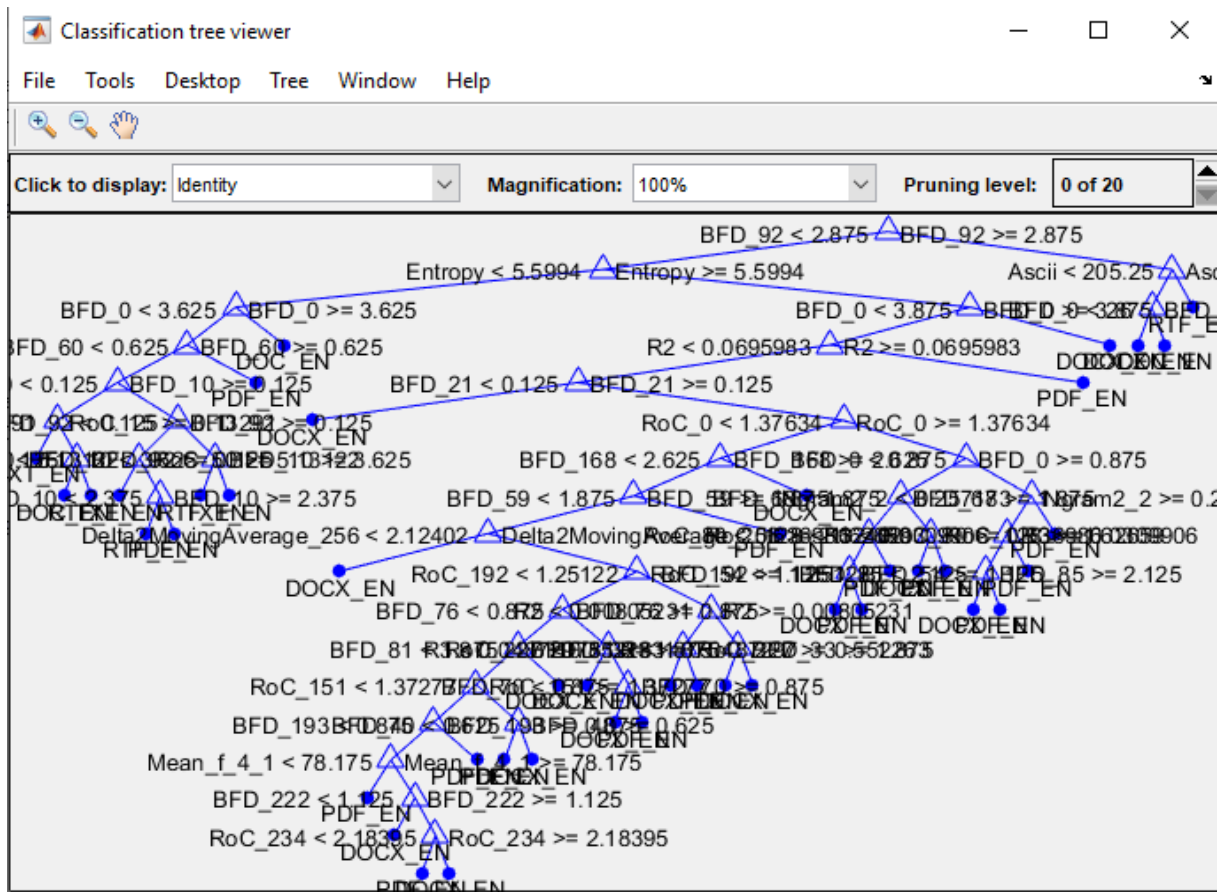

Figure 21: Illustrative example: the trained decision tree.

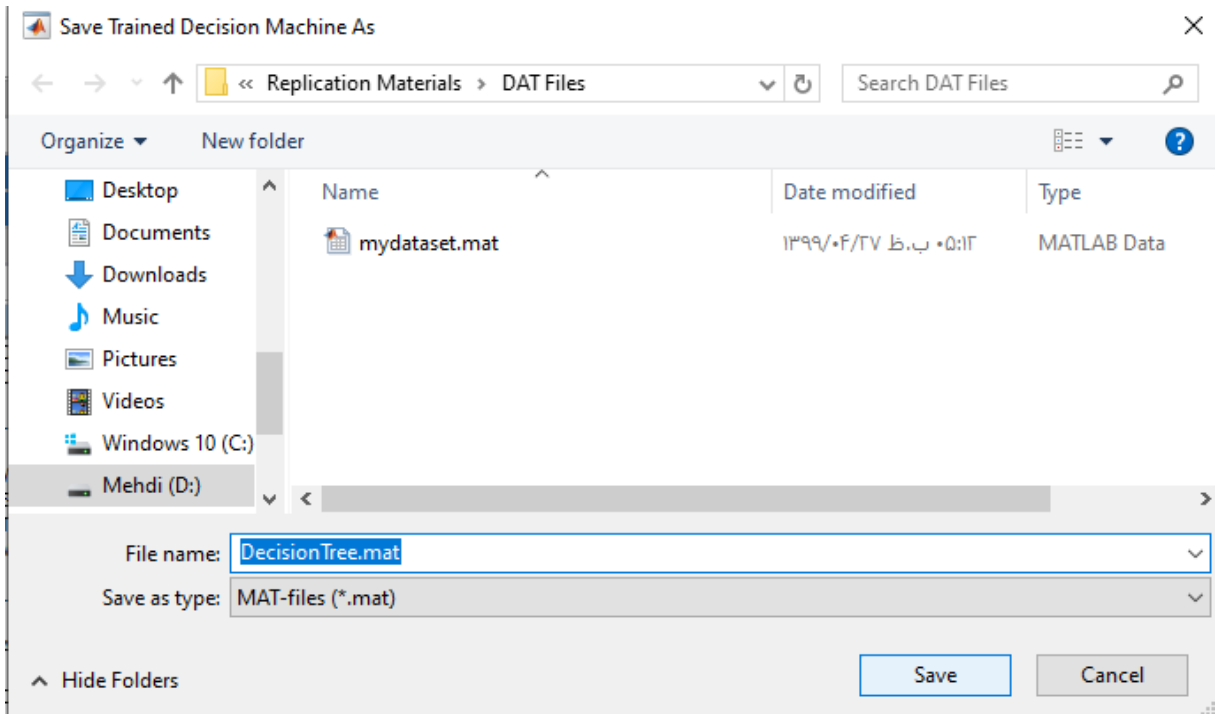

Figure 22: Illustrative example: saving the trained decision machine.

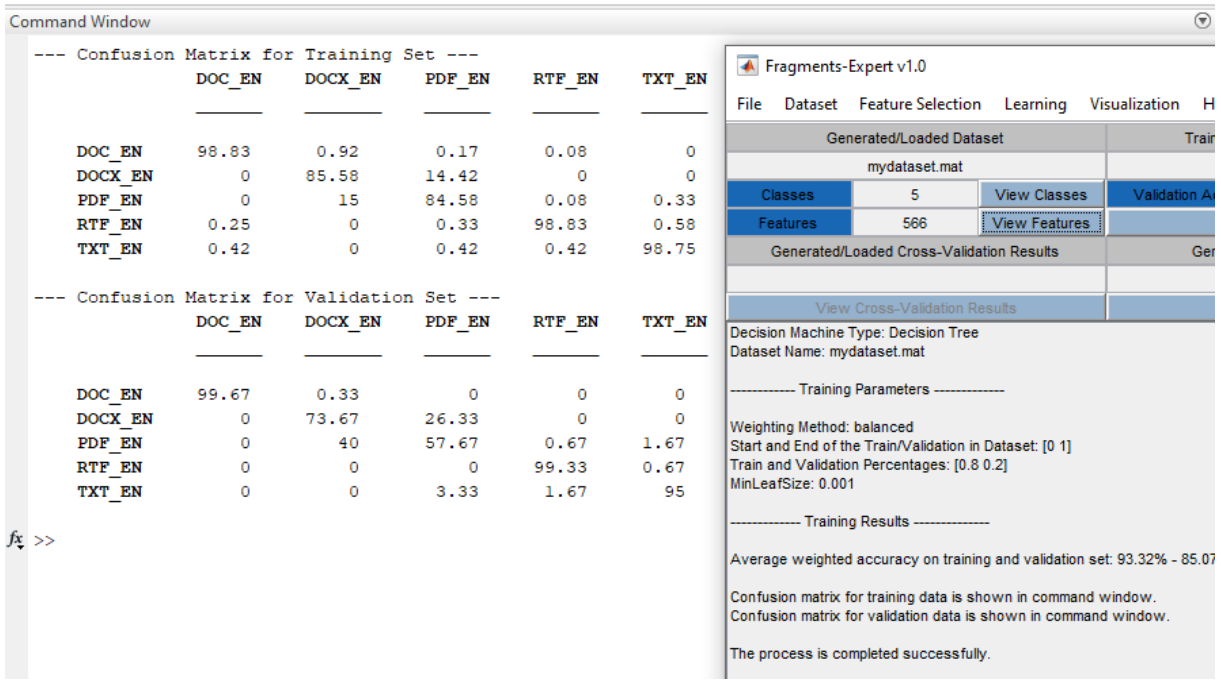

Figure 23: Illustrative example: the training and validation results.

### 5-3 Using Visualization to See the Effect of the Features

As can be seen in Figure 21, $BFD_{92}$, which is equivalent to ASCII character "\", plays an important role in the classification of our textual file formats. Moreover, the ASCII feature, described in Section 2-5, appears near the tree root. This observation indicates the importance of the ASCII feature. As an example, assume that we need to observe PDF and TXT samples in the 2-D space of $BFD_{92}$ and ASCII. This observation helps us to understand the effect of these features in classifying between these two classes.

First, we click on the "Display Samples in Feature Space" submenu in the "Visualization" menu. In the next window, we select two features $BFD_{92}$ and ASCII (see Figure 24) and press the OK button. Then, as shown in Figure 25, we should set the labels for $BFD_{92}$ and ASCII axis. After doing so and pressing the OK button, a list is opened. In this list, we should select the classes that we want to display in our 2-D space. As shown in Figure 26 and Figure 27, we choose PDF and TXT classes. After selecting the second class, i.e. TXT, and pressing the OK button, the list is shown again. In this stage, as we do not need to include another class, we press the cancel button. Then, as shown in Figure 28, we should select the displayed captions for each class in the 2-D plot and press the OK button. After pressing the OK button, a 2-D representation is plotted as shown in Figure 29.

Note that in each stage of class selection, we can select multiple classes. By doing so, we indicate that the samples of the selected classes should be displayed as unified items.

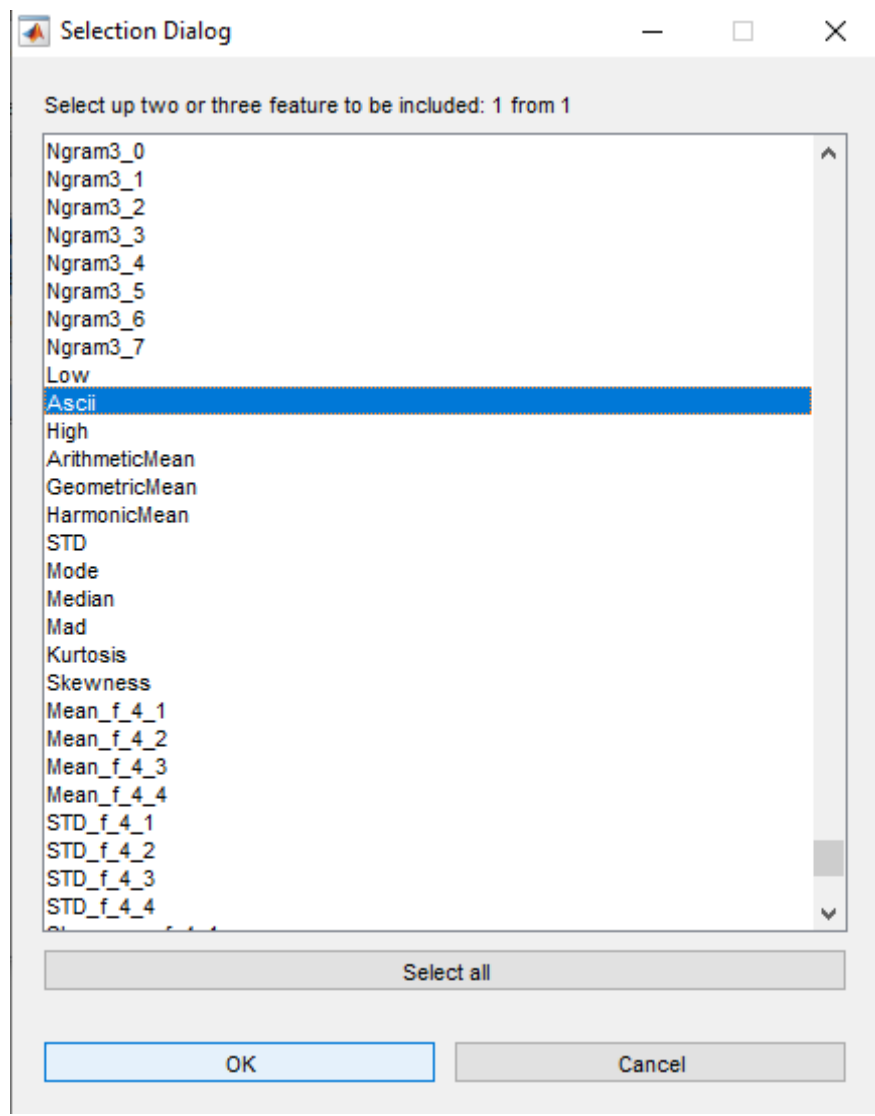

**Figure 24: Illustrative example: selecting $BFD_{92}$ and ASCII features for 2-D representation of samples.**

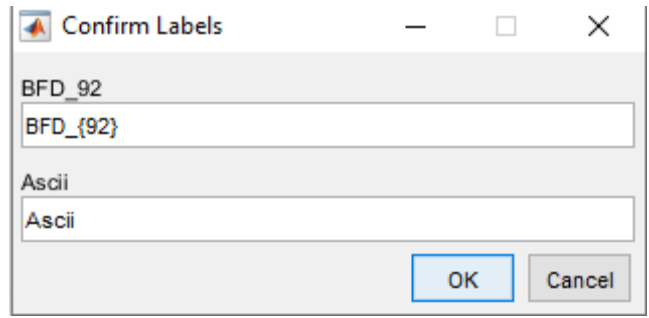

**Figure 25: Illustrative example: choosing the labels for $BFD_{92}$ and ASCII axis.**

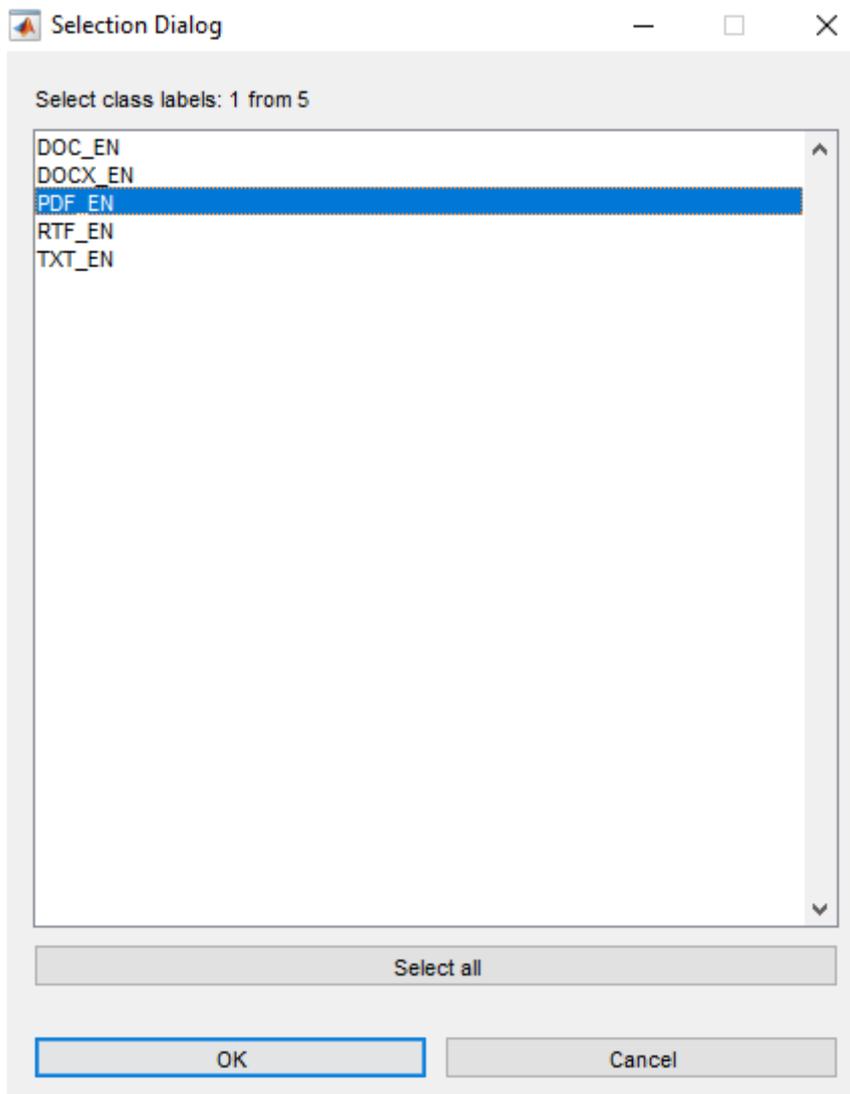

**Figure 26: Illustrative example: choosing class samples PDF as the first class of samples.**

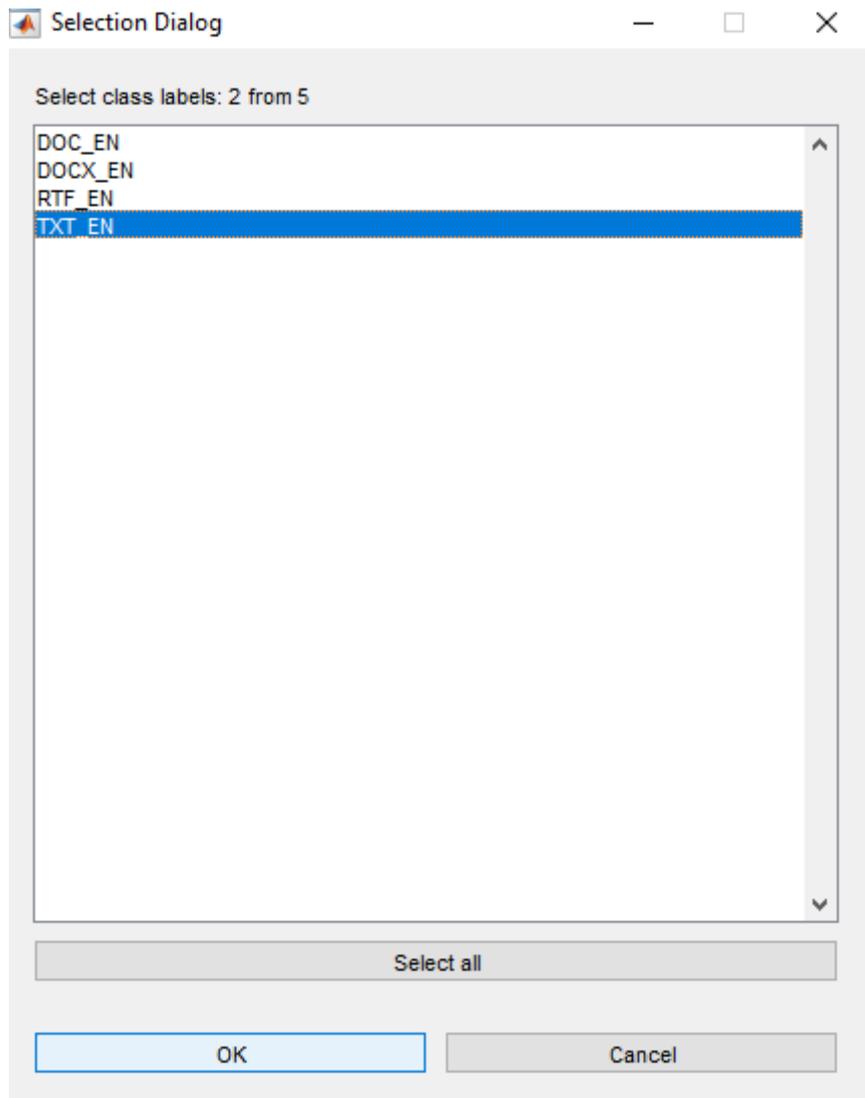

Figure 27: Illustrative example: choosing class samples TXT as the second class of samples.

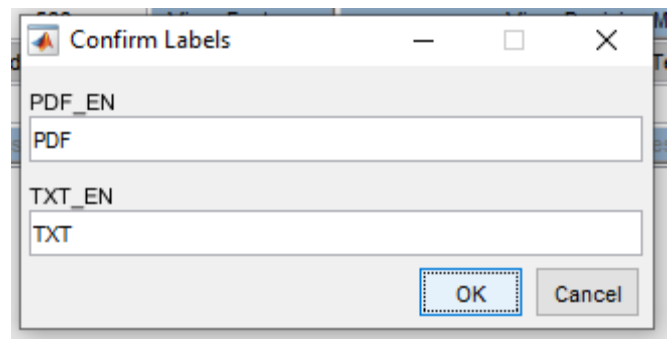

Figure 28: Illustrative example: choosing the displayed labels for two classes.

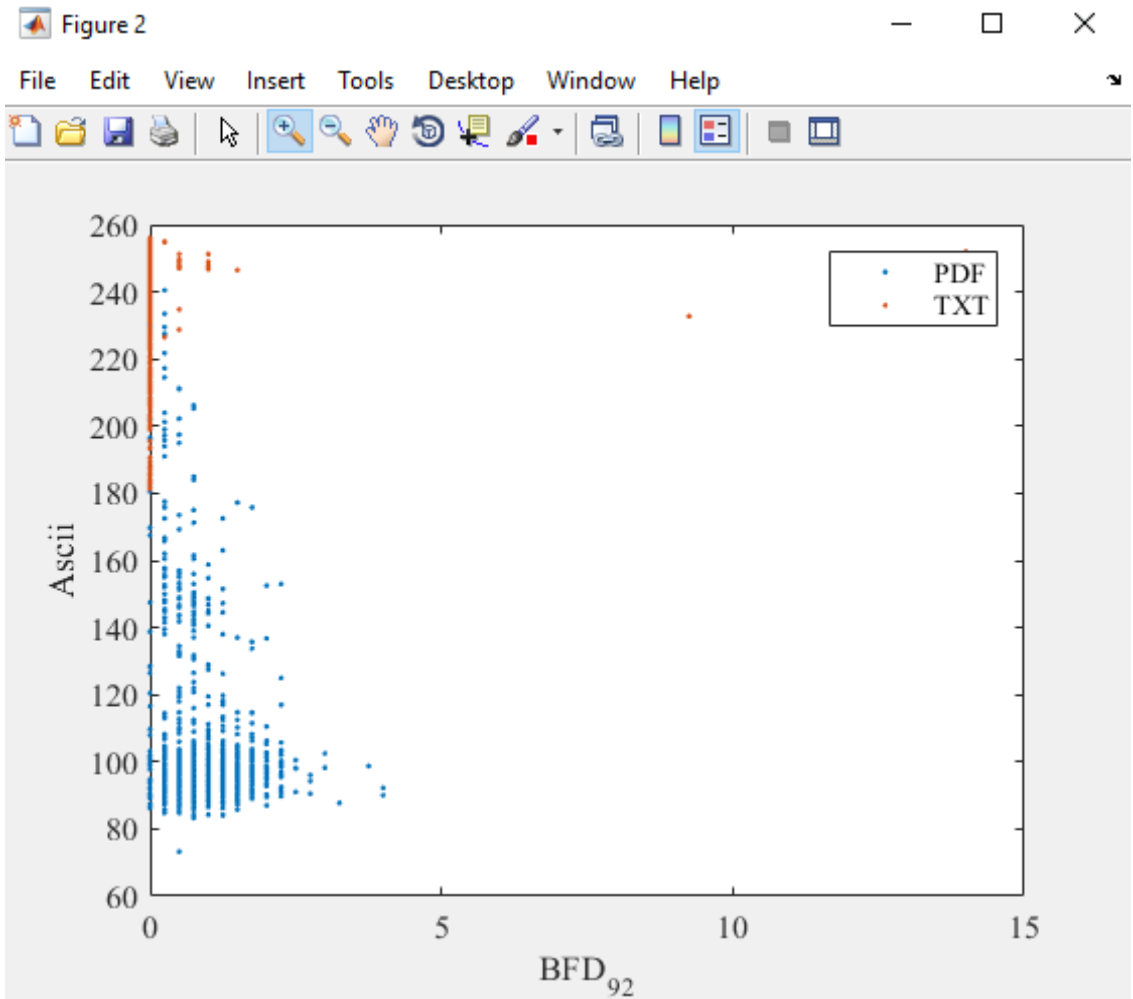

**Figure 29: Illustrative example: 2-D representation of PDF and TXT samples.**

## 5-4   Feature Selection

Many classical machine learning models do not perform well in high-dimensional spaces. When the number of features is too high, the computational complexity of the training process may be unacceptable. Moreover, the risk of overfitting is increased.

In this section, we show how to use the "feature selection" capability of Fragments-Expert. In this case, we use the embedded method of decision-tree learning. To do so, we click on the "Embedded: Decision Tree" submenu in the "Feature Selection" menu. Since, in this case, the toolbox trains a decision tree, we are prompted to input the parameters similar to the parameters described in Section 4-6 for training a decision tree. After setting the parameters and pressing the OK button, the training process starts. After completion of this step, as shown in Figure 30, we see a list of 33 features sorted based on the relative node sizes. We manually select the 15 features with a relative node size greater than 0.05. After pressing the OK button, the new dataset with 15 features is created. Then, as shown in Figure 31, we are prompted to save this new dataset. After saving the dataset, the feature-selected dataset is loaded in the toolbox environment.

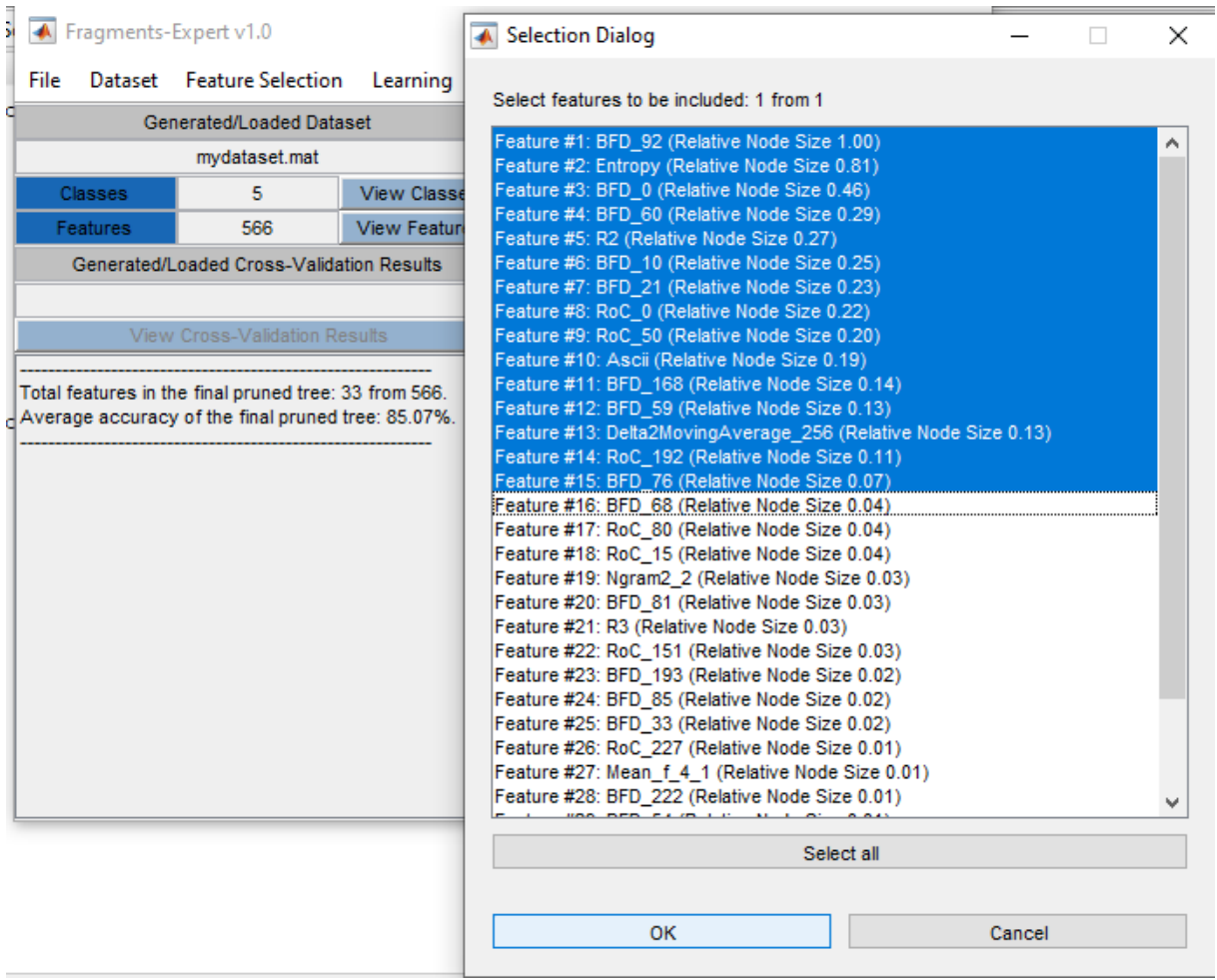

Figure 30: Illustrative example: results of feature selection.

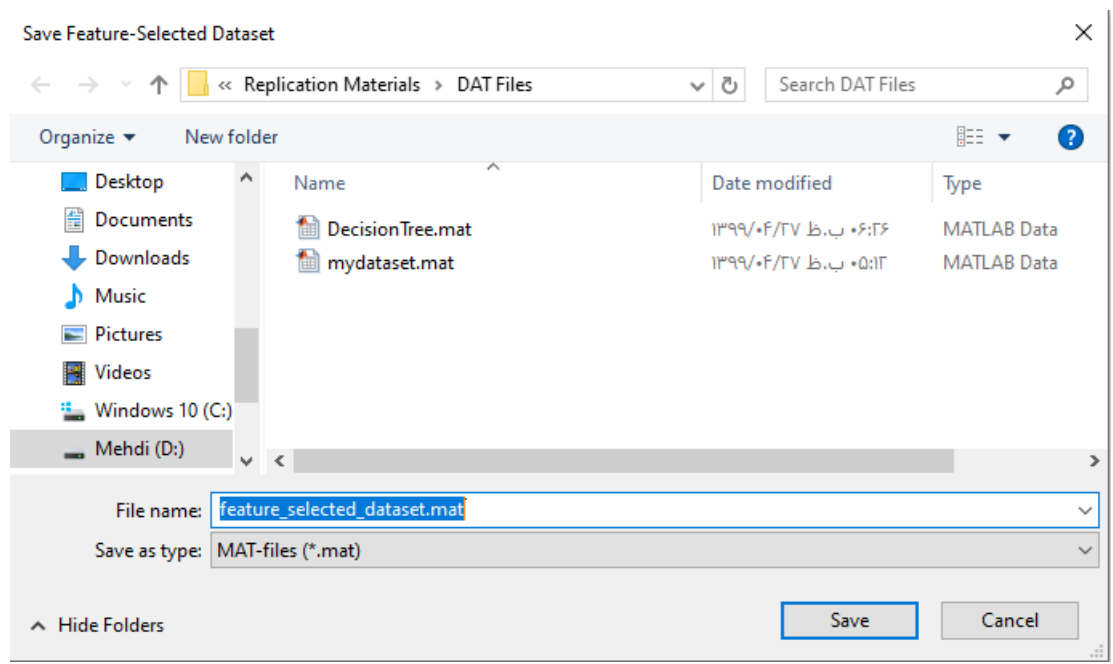

Figure 31: Illustrative example: saving the feature-selected dataset.

## 5-5    Cross-Validation Using Selected Features

In this section, we use cross-validation for assessing the performance of the naïve Bayes method on our feature-selected dataset, which contains 15 selected features. To do so, we click on the "Cross-Validation of Decision Machine" submenu in the "Learning" menu. In the next window, we select "Naive Bayes" and press the OK button.

In the next step, we should set the parameters for cross-validation. As shown in Figure 32, we choose the weighting method "balanced". Also, the start and end of the train/validation/test in the dataset is set equal to [0 1] that indicates using all available samples for train/validation/test. Train and validation percentages are also chosen to be equal to 100% and 0%, respectively. This means that we do not use any validation data in this case. Moreover, we choose z-score for scaling the features. Finally, K=5 is set for K-fold cross-validation.

After setting the parameters and pressing the OK button, the cross-validation process starts and a progress bar is shown until completion of the process. When the cross-validation process is completed, we are prompt to save the results (see Figure 33). After saving the results, the cross-validation results are shown in the main text window of the toolbox. Also, as shown in Figure 34, the cross-validation confusion matrices are shown in the command window of MATLAB. As can be seen in Figure 34, the average performance of the naïve Bayes model on the feature-selected dataset is around 75%.

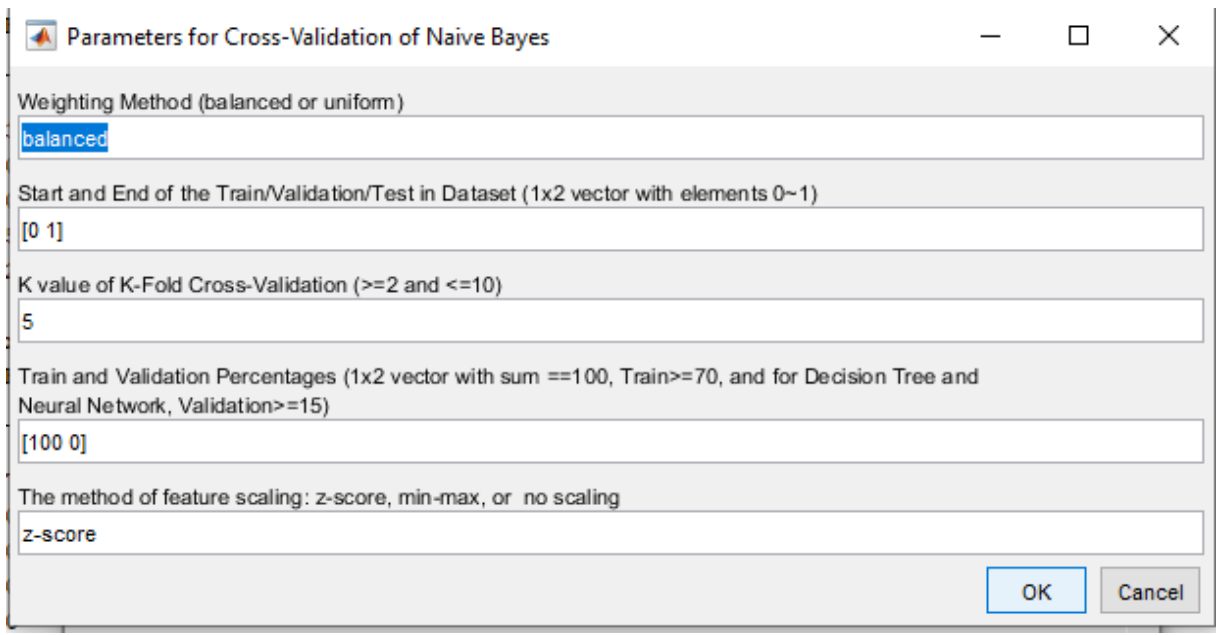

Figure 32: Illustrative example: setting the parameters for naïve Bayes cross-validation.

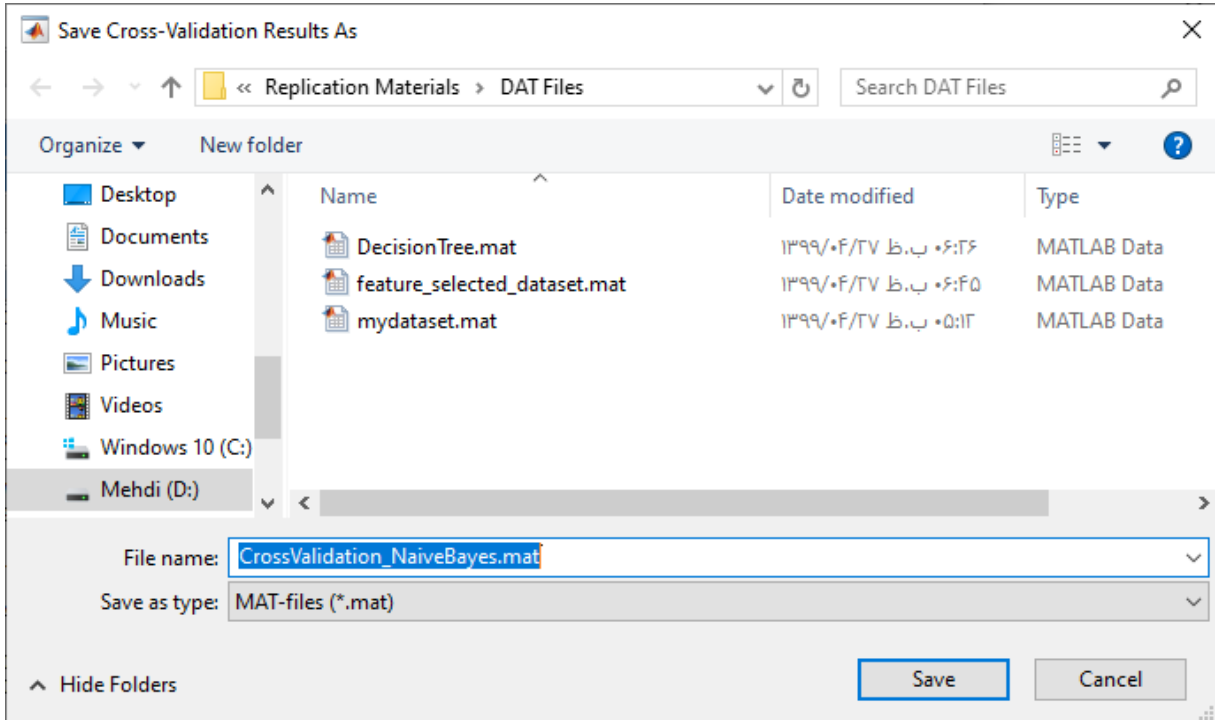

Figure 33: Illustrative example: saving the cross-validation results.

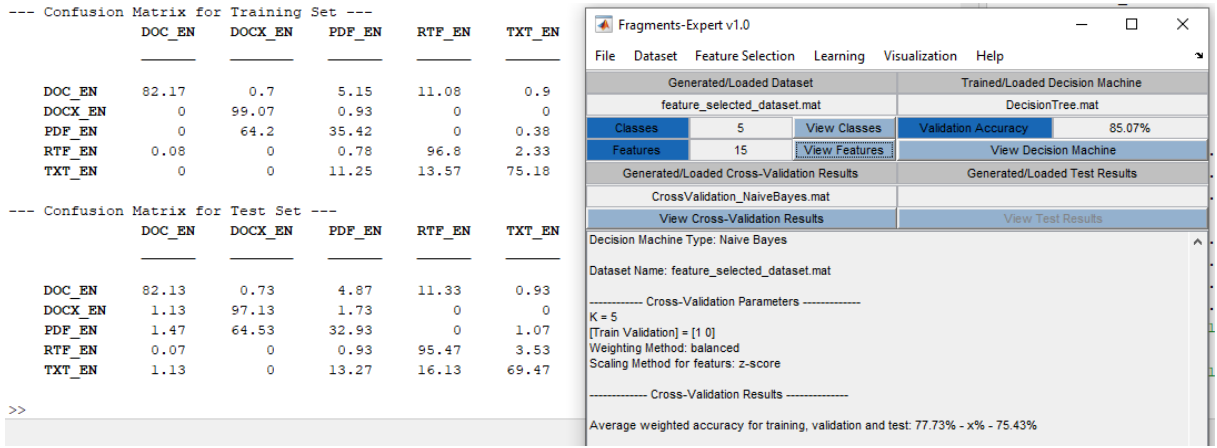

Figure 34: Illustrative example: results of cross-validation.

# 6 Program Availability and Limitations

The version v1.0 of the Fragments-Expert toolbox, which is described in this paper, can be downloaded at https://github.com/mehditeimouri-UT/Fragments-Expert/releases/tag/1.0. To run Fragments-Expert you need to install 64-bit MATLAB R2015b or newer releases for Windows.

There is no specific limit on dataset sizes. However, for large datasets, the speed of toolbox functionalities decreases. The amount of this reduction depends on the hardware specifications of your computer.

The toolbox is not OCTAVE compatible. However, for future works, we plan to make this toolbox compatible with Octave.

# Abbreviations

| | |
|---|---|
| 2-D | 2-Dimensional |
| 3-D | 3-Dimensional |
| BFD | Byte Frequency Distribution |
| BRO | Binary Ratio |
| COTS | Commercial Off-The-Shelf |
| FNF | False Neighbors Fraction |
| GUI | Graphical User Interface |
| $k$-NN | $k$-Nearest Neighbors |
| LDA | Linear Discriminant Analysis |
| LE | Lyapunov Exponent |
| MAD | Mean Absolute Deviation |
| RMS | Root Mean Squared |
| RoC | Rate of Change |
| STD | Standard Deviation |
| SVM | Support Vector Machine |

# Acknowledgments

The authors would like to thank Narges Sadeghi and Fatemeh Delroba, two members of the Information Theory and Coding Laboratory at the University of Tehran, for performing tests to ensure the toolbox functionality and user-friendliness. They also wrote a few functions that are included in the toolbox.